\documentclass[5p,authoryear]{elsarticle}

\usepackage{lineno,hyperref}

\usepackage{flushend}
\usepackage{epsfig}
\usepackage{epstopdf}
\AppendGraphicsExtensions{.tiff}

\usepackage{graphicx}
\usepackage[cmex10]{amsmath}
\usepackage{amssymb}
\usepackage{multirow}
\usepackage{rotating}
\usepackage[table]{xcolor}
\usepackage{threeparttable}
\usepackage{url}

\usepackage{color,soul}
\sethlcolor{green}
\usepackage{algorithmic}
\usepackage{algorithm}
\usepackage{amsthm}

\theoremstyle{definition}

\begin{document}

\begin{frontmatter}
\title{{\em DeepDistance}: A Multi-task Deep Regression Model \\ for Cell Detection in Inverted Microscopy Images}

\author[address1]{Can~Fahrettin~Koyuncu}
\ead{koyuncu@bilkent.edu.tr}

\author[address1]{Gozde~Nur~Gunesli}
\ead{nur.gunesli@bilkent.edu.tr}

\author[address2]{Rengul~Cetin-Atalay}
\ead{rengul@metu.edu.tr}

\author[address1,address3]{Cigdem~Gunduz-Demir\corref{cor1}}
\ead{gunduz@cs.bilkent.edu.tr}
\cortext[cor1]{Corresponding author.}

\address[address1]{Department of Computer Engineering, Bilkent University, Ankara TR-06800, Turkey}
\address[address2]{CanSyL,Graduate School of Informatics, Middle East Technical University, Ankara TR-06800, Turkey }
\address[address3]{Neuroscience Graduate Program, Bilkent University, Ankara TR-06800, Turkey}

\begin{abstract}

This paper presents a new deep regression model, which we call \textit{DeepDistance}, for cell detection in images acquired with inverted microscopy. This model considers cell detection as a task of finding most probable locations that suggest cell centers in an image. It represents this main task with a regression task of learning an \textit{inner distance} metric. However, different than the previously reported regression based methods, the \textit{DeepDistance} model proposes to approach its learning as a multi-task regression problem where multiple tasks are learned by using shared feature representations. To this end, it defines a secondary metric, \textit{normalized outer distance}, to represent a different aspect of the problem and proposes to define its learning as complementary to the main cell detection task. In order to learn these two complementary tasks more effectively, the \textit{DeepDistance} model designs a fully convolutional network (FCN) with a shared encoder path and end-to-end trains this FCN to concurrently learn the tasks in parallel. \textit{DeepDistance} uses the inner distances estimated by this FCN in a detection algorithm to locate individual cells in a given image. For further performance improvement on the main task, this paper also presents an extended version of the \textit{DeepDistance} model. This extended model includes an auxiliary classification task and learns it in parallel to the two regression tasks by sharing feature representations with them. Our experiments on three different human cell lines reveal that the proposed multi-task learning models, the \textit{DeepDistance} model and its extended version, successfully identify cell locations, even for the cell line that was not used in training, and improve the results of the previous deep learning methods. 

\end{abstract}


\end{frontmatter}

\section{Introduction}

Automating the analysis of live cell morphology is critical for high throughput screening as this facilitates fast and reproducible measurements under inverted microscopy. The crucial step of this automation is to correctly identify cell morphology and distribution on culture plates. This requires detecting the cell locations whose difficulty lies along a wide range, from easy to very challenging, depending on visual characteristics of the cells. This step becomes difficult when cells appear in varying colors, brightness, and irregular shapes. The difficulty further increases when they grow in overlayers, and as a result, appear as cell clumps.

\begin{figure}
\centering
\small{
\begin{tabular}{@{~}c@{~}c@{~}c@{~}}
\includegraphics[height=2.75cm]{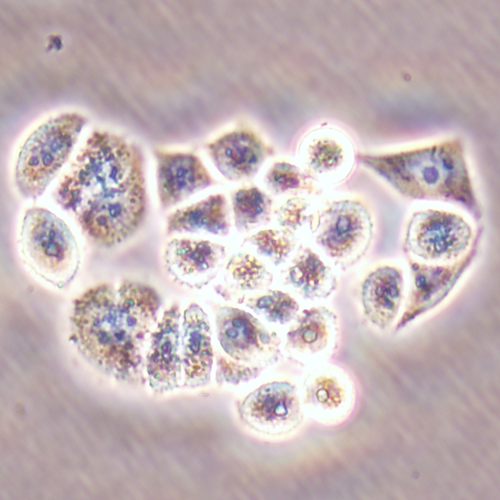} &
\includegraphics[height=2.75cm]{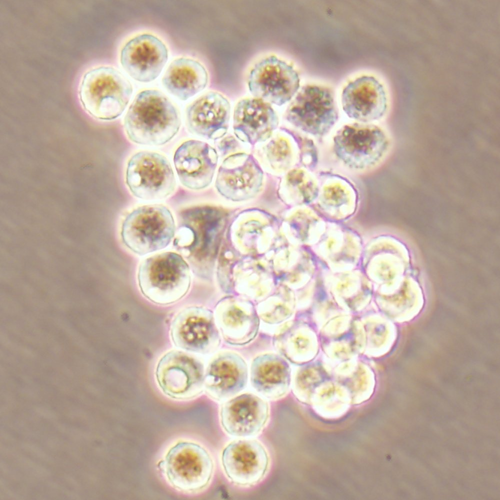} &
\includegraphics[height=2.75cm]{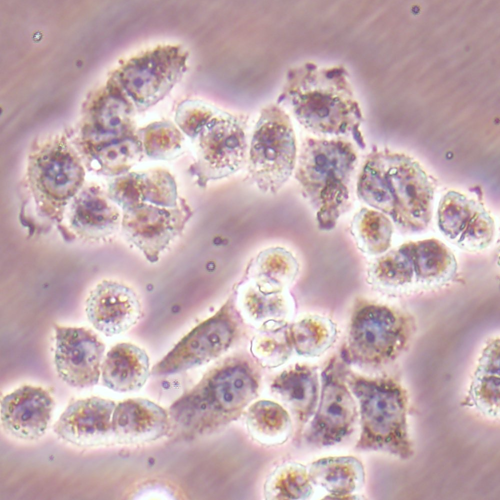} \\
(a) & (b) & (c) \\ 
\end{tabular}
}
\caption{Example subimages from the cell lines used in our experiments: (a) CAMA-1, (b) MDA-MB-453, and (c)  MDA-MB-468 breast cancer cell lines. As seen in these examples, visual characteristics show differences from one cell type to another. Moreover, cells of the same type may appear in different looks. For instance, in (b), there are mostly near-circular cells, which sometimes contain mostly bright pixels but sometimes contain dark pixels inside and bright ones outside. However, in (a) and (c), there are near-circular as well as non-circular cells. For such images, it would not be easy to use a single model to detect cells of all these different looks.}
\label{fig:examples}
\end{figure}

Before the advances in deep learning, the traditional approach for cell detection/segmentation is to employ low-level handcrafted features, reflecting color, edge, and shape characteristics of cells. This approach has given promising results when the features are defined properly, as a good representation of the visual cell characteristics. On the other hand, these characteristics may change from one cell type to another (see Fig.~\ref{fig:examples}) and new features need to be defined to meet the cell characteristics of a new type. Additionally, when there exists heterogeneity in the visual characteristics of the same cell type, using a single model may not be sufficient to detect all cells of this type, particularly for cancer cells which are exploited more in high throughput screening.

Methods based on deep learning, on the other hand, have responded to these issues by having the ability of learning high-level features from image data automatically and reducing the required effort to obtain a generalizable model as a consequence. The majority of the previously reported methods consider cell detection as a classification problem, in which a deep classifier is trained to differentiate cell pixels from those of the background. Since their focus is the classification of cell pixels, these methods treat the pixels taken from the annotated cells in the same way, regardless of their relative positions within the cell, while training their classifier~\citep{xie15b,song17}. On the other hand, the position of a pixel relative to a cell center (or to a cell boundary) may bring about additional information. There exist only a few studies that take this information into account by constructing a regression model~\citep{xie15a,sirinukunwattana16,chen16, xie18a,xie18b}. These studies approach cell detection/segmentation as a \textit{single-task regression problem} where they learn a single distance output for each pixel. On the other hand, it may be difficult to define a single distance metric that comprehends different aspects of the problem and to learn this single distance by a single model. 

In response to these issues, this paper introduces a new multi-task learning framework, which we call \textit{DeepDistance}, for the detection of live cells in inverted microscopy images. This \textit{DeepDistance} framework proposes to concurrently learn two distance metrics for each pixel, where the primary one is learned in regard to the main cell detection task and the secondary distance is defined to stress the variability in morphological cell characteristics and learned for the purpose of increasing the generalization ability of the main task. To this end, this paper constructs a fully convolutional network and end-to-end learns two distance maps at the same time, sharing high-level feature representations at the various layers of this network (layers of its shared encoder path), in the context of multi-task learning. Then, for a given image, it achieves cell detection by generating the primary distance map with the trained network and finding its regional maxima. Our experiments on three different cell lines reveal that this proposed multi-task learning framework successfully identifies cell locations, even for the cell line that was not used in training, and improves the results of the previous deep learning approaches. 

The contributions of this paper are summarized below:
\begin{itemize}
\item It takes advantage of the multi-task learning approach, in which shared feature representations are used to learn multiple tasks at the same time. This is different than the previously reported regression-based cell detection studies, which do not use such shared representations for learning a regression task. The multi-task approach used by our study is known to be successful for many domains, leveraging the contribution of different tasks to the feature representation learning process~\citep{caruana97}. Concurrent learning of two related tasks with shared representations increases the performance of our model, by better helping it avoid local optimal solutions.

\item It defines a distance metric, \textit{normalized outer distance}, that calculates the normalized distance from each cell pixel to the closest cell boundary. As opposed to the \textit{inner distance}, which is calculated with respect to the cell centers and as a result imposes a one-sized circular shape on the cells, this definition does not have shape and size impositions since it uses the boundary annotations. The normalized outer distance better preserves the shape characteristics of the cells whereas the inner distance better suggests the cell centers. Thus, the proposed model defines inner distance estimation as the main task and considers normalized outer distance estimation as complementary to this main task. It learns these two complementary tasks in parallel by forcing them to share feature representations. This improves the performance of each task, and as a result, leads to more successful cell detection.

\item It shows that one can also include an additional classification task to the proposed multi-task regression network to further increase the performance of the main task. To this end, this study implements another version of the proposed framework where the task of cell pixel classification is added as a parallel task to the regression network. This additional task, which is to be concurrently learned with the two distance maps, aims to construct a classification map from the shared features while learning the regression output maps with a minimum error. This additional task is effective to increase the success of the main cell detection task.

\end{itemize}

\section{Related Work}

Traditional cell detection/segmentation studies employ low-level handcrafted features, which are extracted either pixel- or subregion-wise. A large group of pixel-wise studies use intensities to obtain a binary mask by thresholding or clustering~\citep{dima11}. They then use this mask either directly to locate isolated cells or as an input to shape-based methods to split cell clumps. These methods include the use of distance transforms~\citep{jung10}, concavity detection algorithms~\citep{chang13}, and morphological erosion operators~\citep{yang06}. Although it is very common to calculate distance transforms on the obtained binary mask, it is also possible to learn them directly from the handcrafted features~\citep{gao14}. Another group of pixel-wise studies employ pixel gradients to obtain a feature map, on which regional maxima/minima are identified as cell locations. These studies directly use the gradients to define their feature maps~\citep{koyuncu16} or alternatively get pixels voted along their gradient directions and use the votes the pixels take to define their maps~\citep{xing14}. The subregion-wise studies first partition an image into over-segmented subregions (e.g., superpixels), extract handcrafted color, gradient, and shape features from these subregions, and merge them based on their extracted features to obtain cell locations~\citep{genctav12,su13,koyuncu18}. 

To reduce the required efforts for manual feature definition, deep learning based methods learn high-level features from image data. These methods, especially convolutional neural networks, have shown significant success in many tasks related to medical image analysis~\citep{litjens17} also including cell detection/segmentation. Earlier studies train their deep models on small patches cropped around individual pixels to generate an output for each pixel separately. More recently, with the implementations of fully convolutional networks~\citep{long15} and the U-net model~\citep{ronneberger15}, studies have started end-to-end training their models to learn the outputs of all pixels at once. Most of these studies consider cell detection as a classification problem and train a classifier to differentiate cell and background pixels. Then, for a given image, they may obtain a binary mask by estimating the class labels of its pixels with the trained classifier and use this mask as an input to the shape-based methods~\citep{song15,song17}. Alternatively, they may use the class posteriors of the pixels and identify cell locations on this posterior map by either thresholding~\citep{xu16} or clustering~\citep{su15} but mostly finding regional maxima~\citep{ciresan13,dong15,xie15b,sadanandan17}. 

There exist relatively few studies that consider cell detection/segmentation as a regression problem~\citep{kainz15,xie15a,sirinukunwattana16,chen16, xie18a,xie18b}. These studies define their outputs with regard to the Euclidean distance between a pixel and its closest annotated cell center. Most of them calculate this inner distance using only the dot annotations on cell centers without using any boundary (segmentation) information. Thus, they use a threshold to decide pixels for which the distance will be zero (i.e., determine pixels belonging to the background). This thresholding together with the inner distance definition itself impose a one-sized circular shape on the cells, which may not be true for all cell types. These studies approach the learning of this inner distance as a \textit{single-task regression problem}. Different than all these studies, our proposed \textit{DeepDistance} model defines a secondary distance metric that better preserves the morphological characteristics of cells and considers its learning as a complementary task to the main task of cell detection. Additionally, it proposes a \textit{multi-task regression framework} that uses shared feature representations to concurrently learn these two tasks.

There are only a few studies that use a cascaded network architecture for cell detection/segmentation. \cite{ram18} propose a network that sequentially learns a classification mask on an image and then regresses a density map on this classification mask for cell detection in 3D microscopy images. \cite{kechyn18} uses the architecture proposed by~\cite{bai17} for cell segmentation. This architecture is mainly designed to learn an energy function to be used in a watershed algorithm for the purpose of splitting a map of under-segmented components into their corresponding objects. Thus, it requires obtaining the segmentation map of an image beforehand and takes it as an input together with the image. It first learns a gradient map of a distance transform from these inputs and then learns a map of energy levels from the gradients. As opposed to our proposed multi-task framework, both of these networks cascade their tasks in serial and learns them without sharing any representations. On the other hand, our model proposes to learn two regression maps in parallel, in the context of multi-task learning, which forces these tasks to use shared feature representations. The latter approach is known to be more effective to avoid local optima, and as a result, to obtain a more generalizable model~\citep{caruana97}. 

There exists another study that also uses a multi-task framework to detect glands and nuclei in histopathological images. This framework concurrently learns two classification maps, where the first one is the map of gland/nucleus pixels and the other is that of their boundaries. It then combines the two classification maps with a simple fusion function~\citep{chen17}. However, different than our proposed multi-task regression model, this existing study neither considers detection as a regression problem nor learns regression and classification tasks in a single multi-task network. Additionally, its goal is to locate glands/nuclei in fixed and stained histopathological images whereas our aim is to detect cells in inverted microscopy images which is used for high throughput and real-time cell screening.

\section{Methodology}

The proposed \textit{DeepDistance} model relies on formulating cell detection as a regression problem, in which a metric map is estimated to express the degree of pixels suggesting a cell center, and identifying regional maxima on this map as cell locations. This model uses \textit{inner distance} as the primary metric and estimates it by a fully convolutional network (FCN), considering the learning of this metric as the main task in regard to the cell detection problem. On the other hand, as opposed to the previously reported studies, the \textit{DeepDistance} model proposes to approach this learning as a multi-task regression problem, in which multiple regression tasks are learned using shared feature representations. To this end, this model defines a secondary metric, \textit{normalized outer distance}, and considers its learning as a complementary task that represents a different aspect of the problem. The proposed \textit{DeepDistance} model learns this new task in parallel to the main task of cell detection by constructing and end-to-end training an FCN with a shared encoder path, which forces these multiple tasks to learn shared feature representations at various abstraction levels.

The following subsections give the details of the proposed \textit{DeepDistance} model. Section~\ref{sec:fcn} mathematically formulates the distance metrics used to define the tasks. It then gives the architecture of the FCN used for learning these tasks and provides the details of its training. Section~\ref{sec:extended-fcn} discusses how to extend the proposed multi-task regression network to cover an additional task(s), by giving the details of another version of the proposed model where cell pixel classification is considered as the additional task. Finally, Section~\ref{sec:cell-detection} presents the detection algorithm that uses inner distances estimated by the FCN to locate individual cells in a given image.

\subsection{Multi-task FCN for Distance Learning}
\label{sec:fcn}

The proposed \textit{DeepDistance} model uses two distance metrics for each pixel $q$. The first one is \textit{inner distance} $d_{inner}(q)$ that is calculated similar to the previous studies. Its learning is considered as the main task; cell locations are detected on the estimated inner distance map of the pixels (see Section~\ref{sec:cell-detection}). The second metric is \textit{normalized outer distance} $d_{outer}(q)$ that is defined by this current study in order to better quantify morphological cell characteristics. Learning of this outer distance is considered as a complementary task, which is used to improve the performance of the main task. 

These distances are defined in Eqns.~\ref{eqn:innerdistance} and~\ref{eqn:outerdistance}, respectively, when cell annotations are provided. The annotations are, of course, not available for images whose cells are supposed to be automatically detected. Thus, our \textit{DeepDistance} model proposes to estimate the distances by an FCN that will be trained on the pixels of annotated images. Note that in our experiments, we train this FCN using the annotations of only four training and two validation images. 

Let ${\cal A} = \{a_i\}$ be the set of annotated cells in an image, ${\cal P}(a_i) = \{p_{ik}\}$ be the set of pixels belonging to an annotated cell $a_i$, ${\cal B}(a_i) = \{b_{ik}\}$ be the set of its boundary pixels, and ${\cal C}(a_i)$ be its centroid pixel. For pixel $q$,
\begin{equation}
d_{inner}(q) = \left\{
\begin{array}{@{~}l@{~~}l}
\dfrac{1}{1 + \alpha \min\limits_{a_{i} \in {\cal A}}\lVert q - {\cal C}(a_i) \rVert^2}		& \mbox{if}~q \in {\cal P}(a_i) \vspace{0.1in} \\
\mbox{0}		& \mbox{if}~q \in  \mbox{\small{background}}\\
\end{array}
\right.
\label{eqn:innerdistance}
\end{equation}
\begin{equation}
d_{outer}(q) = \left\{
\begin{array}{@{~}l@{~~}l}
\dfrac{\min\limits_{b_{ik} \in {\cal B}(a_i)}\lVert q - b_{ik} \rVert^2}{\max\limits_{r \in {\cal P}(a_i)} \min\limits_{b_{ik} \in {\cal B}(a_i)}\lVert r - b_{ik} \rVert^2}		& \mbox{if}~q \in {\cal P}(a_i) \vspace{0.1in} \\
\mbox{0}		& \mbox{if}~q \in  \mbox{\small{background}}\\
\end{array}
\right.
\label{eqn:outerdistance}
\end{equation}
where $\alpha$ in Eqn.~\ref{eqn:innerdistance} is the decay ratio that is empirically selected as 0.1, similar to the previous studies. The denominator in Eqn.~\ref{eqn:outerdistance} corresponds to the maximum distance in the annotated cell $a_i$, which is used as a normalization factor. This normalization is effective to obtain similar distances for cells of different sizes, which will drive the FCN to make better generalizations regardless of the cell size. 

\begin{figure}
\centering
\small{
\begin{tabular}{cc}
\includegraphics[width=0.35\columnwidth]{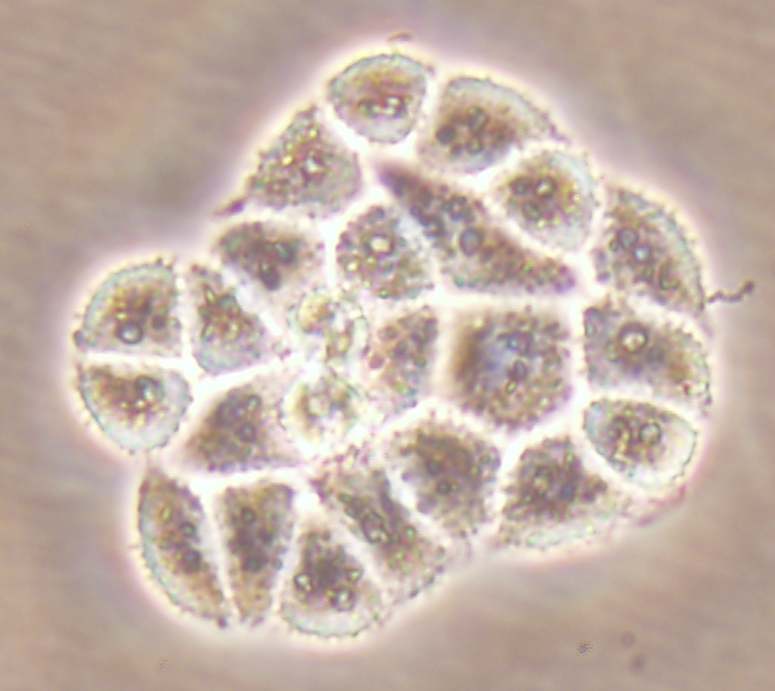} &
\includegraphics[width=0.35\columnwidth]{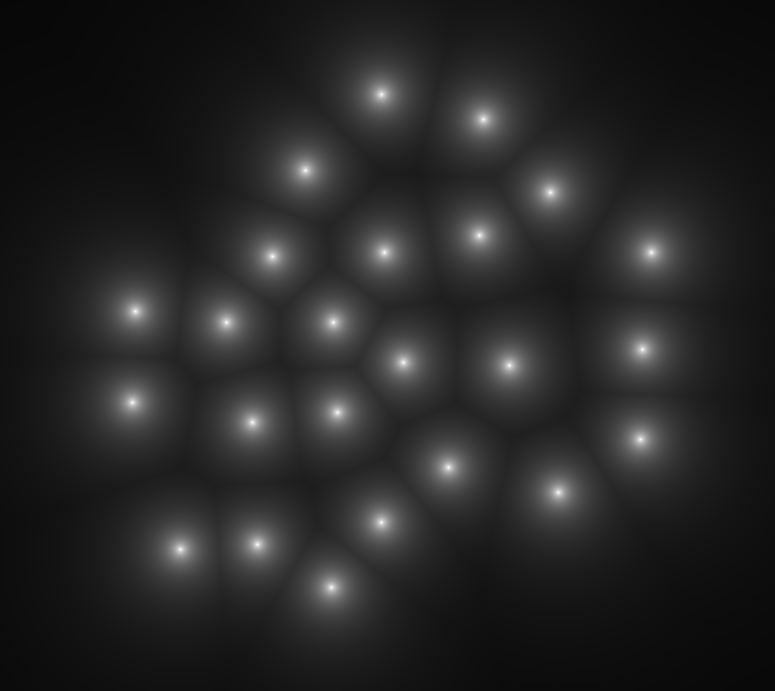} \\
(a) & (b)\\ 
\includegraphics[width=0.35\columnwidth]{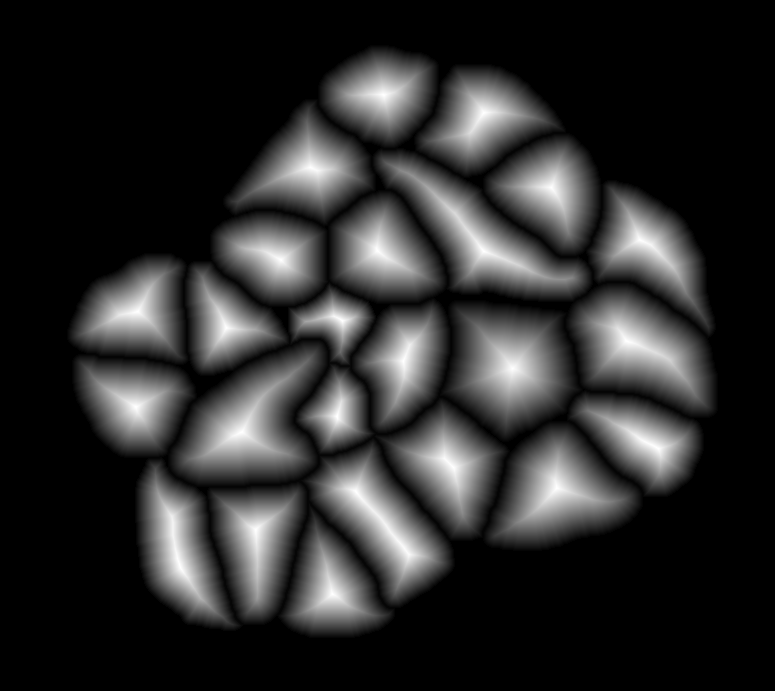} &
\includegraphics[width=0.35\columnwidth]{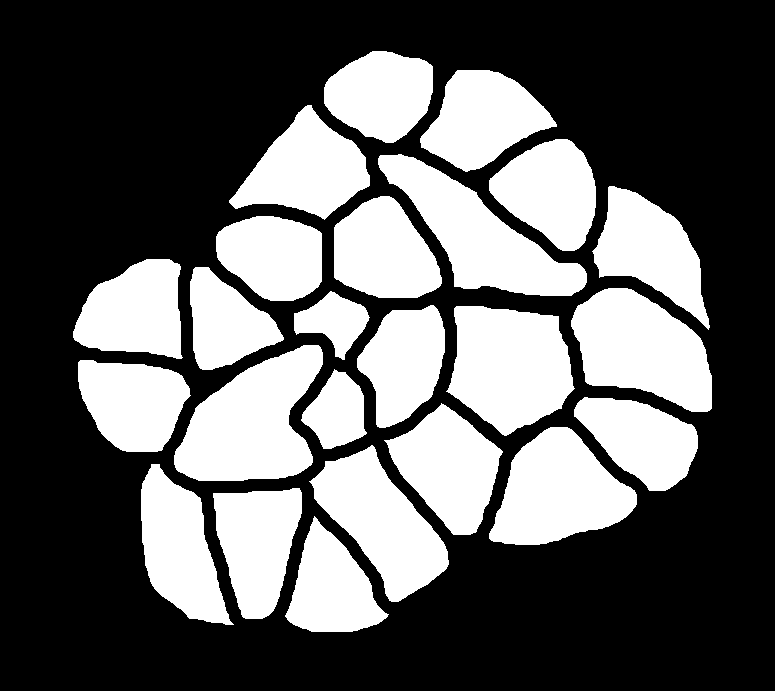} \\
(c) & (d)\\ 
\end{tabular}
}
\caption{(a) Original subimage, (b) inner distance map that uses distances from pixels to their closest cell centers, (c) normalized outer distance map that uses distances from cell pixels to their closest boundary annotations, and (d) cell pixel annotations.}
\label{fig:dist}
\end{figure}

For an example subimage given in Fig.~\ref{fig:dist}a, these distance definitions are illustrated in Figs.~\ref{fig:dist}b and~\ref{fig:dist}c, respectively. The inner distance definition well indicates the cell centers since it uses the Euclidean distances from pixels to their closest cell centers. However, as it uses the centers as the reference point, the distance decrease from a center to its boundaries is the same for all directions and for all cells. Thus, when it is used alone, this definition imposes a circular and one-sized shape on the cells, as also seen in Fig.~\ref{fig:dist}b. On the contrary, since the normalized outer distance is calculated with a reference to a cell boundary, this decrease may differ from one direction to another as well as from one cell to another, depending on the shape and size of the cell. Thus, it better preserves the morphological characteristics of cells, as seen in Fig.~\ref{fig:dist}c.

\subsubsection{Network Architecture}
\begin{figure*}
\centering
\includegraphics[width=2\columnwidth]{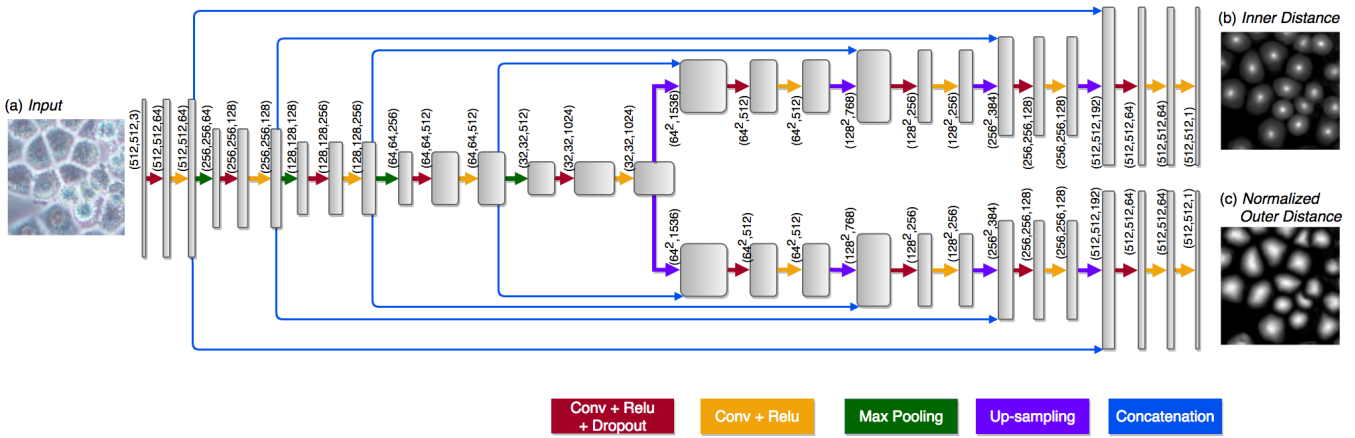} \\
\caption{Architecture of the fully convolutional network (FCN) used for multi-task regression of two distance maps, along with an example tile (a) as the input and the predicted (b) \textit{inner distance} map and (c) \textit{normalized outer distance} map as the outputs. Note that the tile used in this figure is not a part of the training set used in our experiments. Each box represents a multi-channel feature map with its dimensions and number of channels being indicated in order on the left side of the box. Each arrow corresponds to an operation which is distinguishable by its color.}
\label{fig:fcn-architecture}
\end{figure*}

For multi-task learning of these two distance maps, our \textit{DeepDistance} model constructs an FCN architecture consisting of a shared encoder path and two decoder paths (see Fig.~\ref{fig:fcn-architecture}). The encoder path is shared by the two tasks to extract shared feature representations from an RGB image, whose pixels are normalized across the image, at various abstraction layers. The two decoder paths, with symmetric connections to the features in the encoder path (shown with concatenation operators in the figure), are used to separately construct the distance maps from these extracted shared features. This architecture has the convolution layers with $3 \times 3$ filters and uses the rectified linear unit (ReLU) activation function. Its pooling/upsampling layers use $2 \times 2$ filters. The number of the layers and the number of the feature maps used in each convolution layer are depicted in Fig.~\ref{fig:fcn-architecture}. Note that these numbers are selected by inspiring with the U-net model~\citep{ronneberger15}. The original U-net model has a single decoder path designed for single-task learning. On the contrary, the \textit{DeepDistance} model has two decoder paths, with symmetric connections to the shared features, for multi-task learning of the two distance maps. 

This FCN is end-to-end trained on $512 \times 512$ tiles cropped out of the training images. This tile size makes maximum use of the memory on the GPU that we used (GeForce GTX 1080 Ti) when the batch size is selected as 1. The tiles are cropped by a sliding window approach with an increment of 256 pixels. The selection of this increment size ensures that regions stay on the borders of one tile will be close to the central area of another tile. 

\subsubsection{Network Training}
The FCN is implemented in Python using the Keras deep learning library. It is trained from scratch with the backpropagation algorithm that uses the mean squared error as its loss function. It follows an early stopping approach based on the loss calculated for the tiles cropped out of the validation images. The contributions of both tasks to the loss function are the unit weight. The batch size is 1 and the drop-out factor is 0.2. The learning rate and the momentum value are adaptively adjusted using the AdaDelta optimizer~\citep{zeiler2012}. The source codes of this implementation are available at http://www.cs.bilkent.edu. tr/$\scriptstyle\mathtt{\sim}$gunduz/downloads/DeepDistance.

\subsection{Extending the FCN for Additional Tasks}
\label{sec:extended-fcn}

The proposed \textit{DeepDistance} model considers cell detection as a multi-task regression problem that estimates two distance maps from the RGB image, one for formulating the main task of cell detection and the other as an auxiliary task with the motivation of more effectively learning the main task. The FCN architecture given in the previous section is designed to learn these two regression tasks at the same time. This section discusses how this model can be extended to cover more auxiliary tasks, concurrent learning of which may further increase the performance of the main task. For this purpose, this section implements an extended version of the \textit{DeepDistance} model that comprises an additional task of cell pixel classification. This additional task aims to construct a classification map (as shown in Fig.~\ref{fig:dist}d) from the shared features of the encoder path\footnote{To take overlapping cells apart, and hence to obtain an improved map, cell boundaries are widen and subtracted from the classification map. This improved map is also used in the comparison methods to make fair comparisons.}. Note that here, instead of defining another regression problem as the additional task, we use a classification problem in order to demonstrate that the model can easily be extended to cover the auxiliary tasks related with regression as well as classification. 

The extended version of the \textit{DeepDistance} model uses the FCN that has still one shared encoder path but one extra decoder path, defined for the new classification task. The architecture of this new decoder path is the same with those of the two decoder paths, defined for regressing the distance maps, except that its last convolution layer uses the sigmoid function instead of ReLU. Other than this, it has the same convolution and upsampling layers and uses the same symmetric connections to the features in the encoder path (uses the same concatenation operators). Training of this extended FCN follows the same procedure explained in the previous section, with only a difference of loss calculation. This extended FCN still uses the mean squared error as its loss function and the regression tasks still equally contribute to this loss with a unit weight, but the loss contribution weight of the new classification task is 0.1. The rationality behind using a reduced weight is as follows. Both of the distance outputs, calculated as defined in Eqns.~\ref{eqn:innerdistance} and~\ref{eqn:outerdistance}, are in the range between 0 and 1. However, these distances reach the maximum value of 1 for only a few cell pixels whereas they yield much smaller values for the rest of them. On the other hand, the output of the classification task is always 1 for the cell pixels, which results in calculating a larger mean squared error for this task. Since all tasks are learned at the same time by sharing the same features, to avoid creating an unfair bias towards the learning of the classification task, we reduce its loss contribution weight to 0.1. 

\subsection{Cell Detection}
\label{sec:cell-detection}

The last step of the \textit{DeepDistance} model is to detect cells in an unannotated image. For that, this step feeds the tiles cropped out of the image to the trained FCN and identifies cell locations on the input distance maps estimated by this FCN. Since pixels belonging to a cell center are expected to have higher estimated values, the \textit{DeepDistance} model identifies regional maxima on the inner distance maps as the cell centers. In order to suppress possible noise in the estimated distance maps, the model applies the h-maxima transform beforehand and suppresses the maxima whose height is less than the value of $h$.

This step may result in poor estimations for the regions close to the tile edges. As a solution to this problem, our model estimates the inner distance maps for overlapping tiles and then averages all distances calculated for the same pixel. The overlapping tiles are obtained by sliding a window over the image with an increment of 64 pixels. Considering the $512 \times 512$ tile size used by the FCN, this increment size is small enough to ensure that the regions close to the edges of one tile will be close to the central region of some others. It is also large enough to cause only negligible speed-down in the computational time. 

\section{Experiments}
\label{sec:experiments}

\subsection{Datasets}

We test our \textit{DeepDistance} model on three datasets, each of which consists of live cell images of a different  cell line. They are the CAMA-1, MDA-MB-453, and MDA-MB-468 human breast cancer cell lines. The images in all datasets were acquired at $20 \times$ magnification and $3096 \times 4140$ pixel resolution. An example image from each dataset is shown in Fig.~\ref{fig:examples}. As seen in this figure, cells might be visually different within and across different cell lines. 

Three images are randomly selected from each of the CAMA-1 and MDA-MB-453 cell lines and are used for training the FCN as well as for selecting the $h$ parameter of the cell detection step. While training the FCN, the tiles cropped out of four of these six images are used to learn the weights of the FCN and those of the remaining two are used as validation tiles for early stopping. The cells in the rest of the images in these two cell lines are used for testing. In our experiments, these test cells, which belong to CAMA-1 and MDA-MB-453, are considered as dependent test samples since other images/cells of the same cell lines are used for training. To assess the success of our model on an unseen cell line, none of the images of MDA-MB-468 are used for training the FCN or for parameter selection and all of them are used just for testing. Thus, the cells of this MDA-MB-468 cell line are considered as independent test samples. For each cell line, the number of images and the number of cells in the training, validation, and test sets are presented in Table~\ref{table:datasets}.
\begin{table}[t]
\caption{For each cell line, the number of images and the number of cells in its training, validation, and test sets.}
\label{table:datasets}
\centering
\small{
\begin{tabular}{|@{~}l@{~}|c@{~}|@{~}c@{~}|@{~}c@{~}|@{~}c@{~}|@{~}c@{~}|@{~}c@{~}|}
\hline
& \multicolumn{2}{c|@{~}}{Training} &  \multicolumn{2}{c@{~}|@{~}}{Validation} &\multicolumn{2}{c@{~}|}{Test} \\ \cline{2-7}
				& Image & Cell & Image & Cell & Image & Cell \\ \hline
CAMA-1 & 2 & 752 & 1 & 84 & 6 & 1253 \\ \hline
MDA-MB-453  & 2 & 522 & 1 & 137 & 4 & 776 \\ \hline
MDA-MB-468 & - & - & - & - & 8 & 1668 \\ \hline
\textbf{\emph{Total}} & 4 & 1274 & 2 & 221 & 18 & 3697 \\ \hline
\end{tabular}
}
\end{table}

The cells are annotated by putting markers to their approximate centers. These markers are used in the cell-level evaluation of our \textit{DeepDistance} model as well as the comparison methods (Section~\ref{sec:evaluation}). In addition to these markers, training cells are also annotated by delineating their precise boundaries since the definition of the normalized outer distance to be learned by the FCN requires knowing the cell boundaries. It is worth to noting that although annotating boundaries requires more effort, our model needs the boundary annotations only for the training and validation images (only for six images in our experiments).

\subsection{Evaluation}
\label{sec:evaluation}

Each method is quantitatively evaluated on the test cells regarding the f-score metric. This metric is calculated at the cell-level, considering the number of one-to-one matches between the annotated markers and the detected cells (detected regional maxima). For that, each annotated marker is matched to every detected cell if the distance between this marker and the centroid of the detected cell is less than a distance threshold. Similarly, each detected cell is matched to every annotated marker if their distance is less than the same threshold. Afterwards, a detected cell $C$ is considered as one-to-one match (true positive), if it matches with only one marker and this marker matches with only the cell $C$. Considering the image resolution and the cell sizes, this threshold is selected as 30. Then, the precision and recall metrics are obtained on these one-to-one matches, and the f-score is calculated as the harmonic mean of these two metrics. 

\subsection{Parameter Selection}

The \textit{DeepDistance} model has one external parameter: The $h$ value used by the h-maxima transform in the cell detection step to suppress possible noise in the estimated inner distance map. The value of this parameter is selected on the six images belonging to the training and validation sets of the CAMA-1 and MDA-MB-453 cell lines. For that, the following values of $h = \{0.1, 0.2, 0.3, 0.4, 0.5\}$ are considered and the one that yields the highest f-score metric for the cells in these six images is selected. This procedure selects $h = 0.2$ for both the \textit{DeepDistance} model and its extended version. Note that the parameter values are selected similarly for the comparison methods.

\begin{table*}[t]
\centering
\caption{F-score metrics obtained on the test sets by the proposed \textit{DeepDistance} models and the comparison methods.}
\label{table:results}
\small{
\begin{tabular}{|l|c|c|c|}
\hline
 & \multicolumn{2}{c|}{\cellcolor[gray]{0.75} Dependent test samples} & \cellcolor[gray]{0.75} Independent test samples  \\ \cline{2-4}
 & ~~~~~~CAMA-1~~~~~~ & ~~MDA-MB-453~~ & MDA-MB-468 \\ \hline
\bf{DeepDistance} 				& 90.90 & 91.79 & 86.04 \\ \hline 
\bf{DeepDistance (extended)} 		& 91.25 & 92.19 & 86.35 \\ \hline 
SingleInner 					& 87.40 & 89.85 & 82.30 \\ \hline 
SingleOuter 					& 90.80 & 90.78 & 81.43 \\ \hline 
SingleClassification 				& 83.66 & 86.52 & 61.42 \\ \hline 
CascadedClassificationInner 		& 84.49 & 87.76 & 71.43 \\ \hline 
MultiClassificationBoundary	 	& 90.03 & 91.47 & 77.37 \\ \hline 
\end{tabular}
}
\end{table*}

\begin{figure*}[!t]
\centering
\footnotesize{
\begin{tabular}{@{~}c@{~}c@{~}c@{~}c@{~}c@{~}c@{~}c@{~}c@{~}c@{~}}
%
\begin{sideways}\bf{~~~Dependent}\end{sideways} &
\includegraphics[width =0.234\columnwidth]{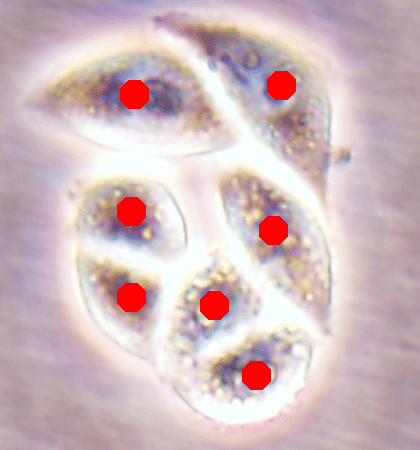} & 
\includegraphics[width =0.234\columnwidth]{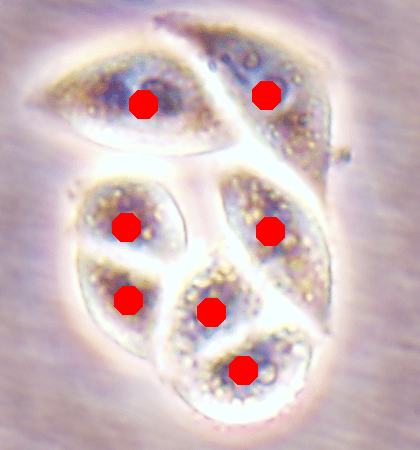} & 
\includegraphics[width =0.234\columnwidth]{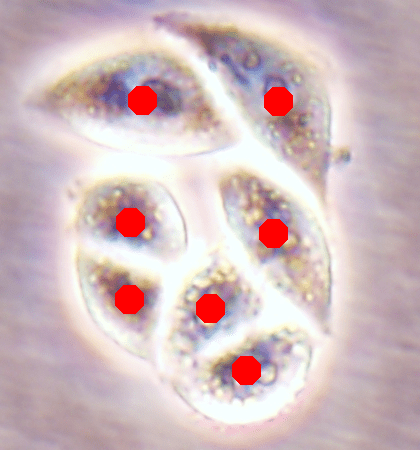} & 
\includegraphics[width =0.234\columnwidth]{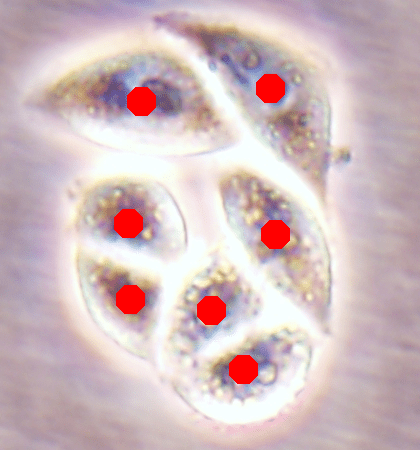} & 
\includegraphics[width =0.234\columnwidth]{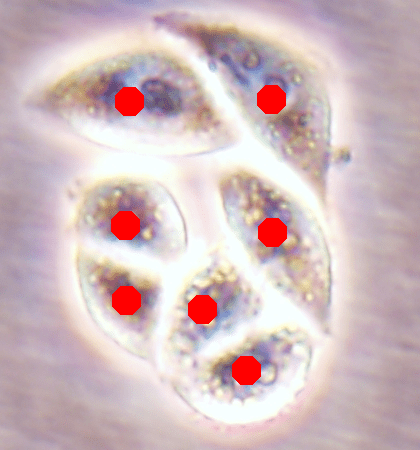} & 
\includegraphics[width =0.234\columnwidth]{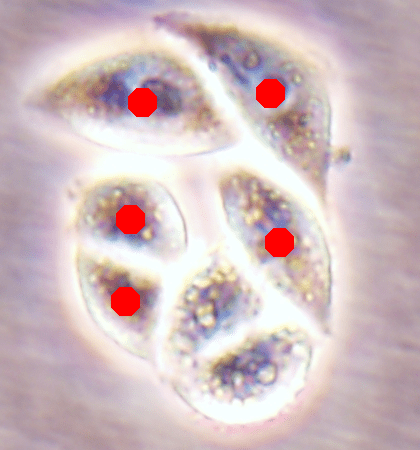} & 
\includegraphics[width =0.234\columnwidth]{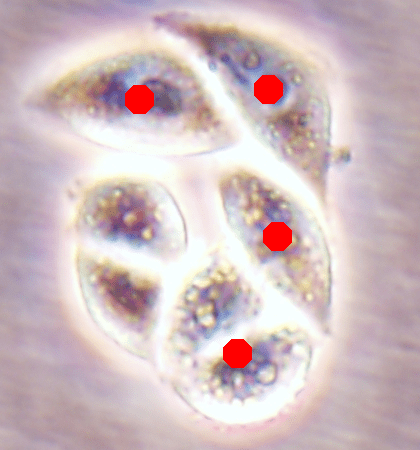} & 
\includegraphics[width =0.234\columnwidth]{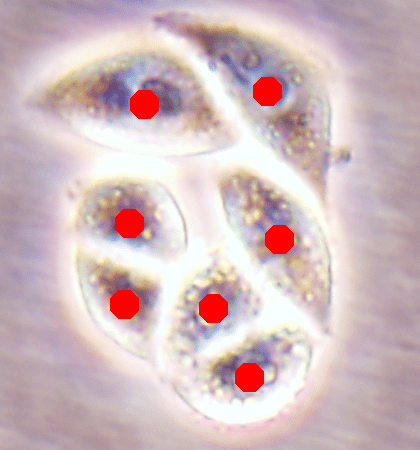} \\
\begin{sideways}\bf{~~~Dependent}\end{sideways} &
\includegraphics[width =0.234\columnwidth]{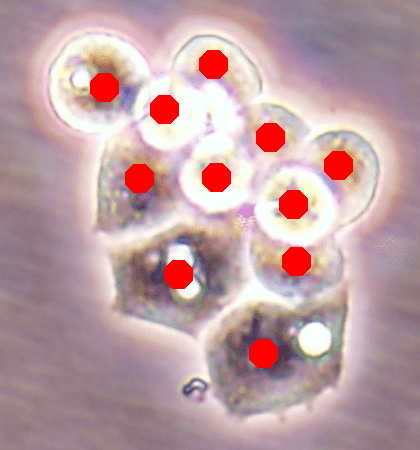} & 
\includegraphics[width =0.234\columnwidth]{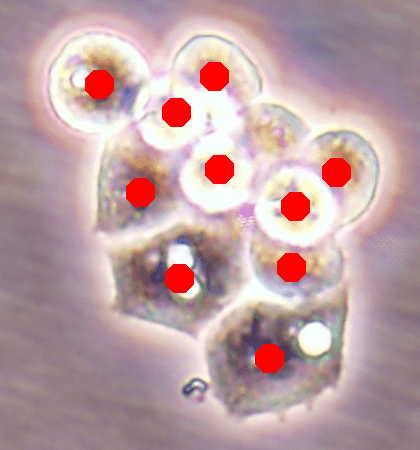} & 
\includegraphics[width =0.234\columnwidth]{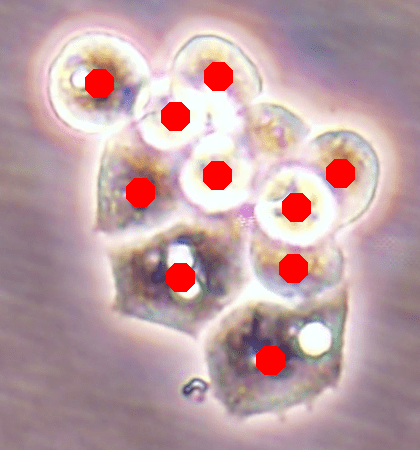} & 
\includegraphics[width =0.234\columnwidth]{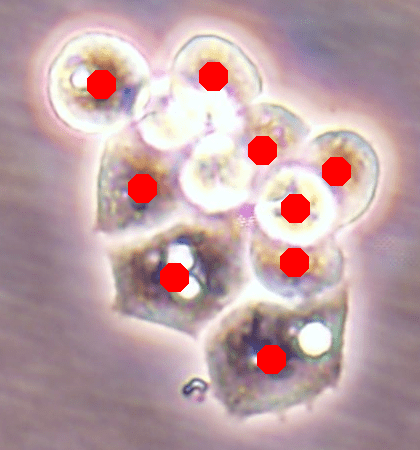} & 
\includegraphics[width =0.234\columnwidth]{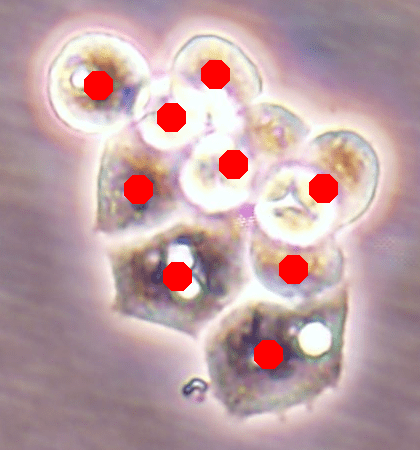} & 
\includegraphics[width =0.234\columnwidth]{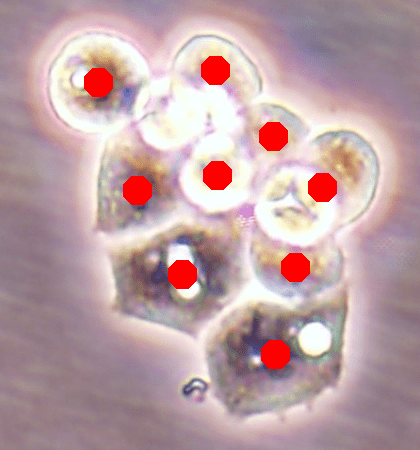} & 
\includegraphics[width =0.234\columnwidth]{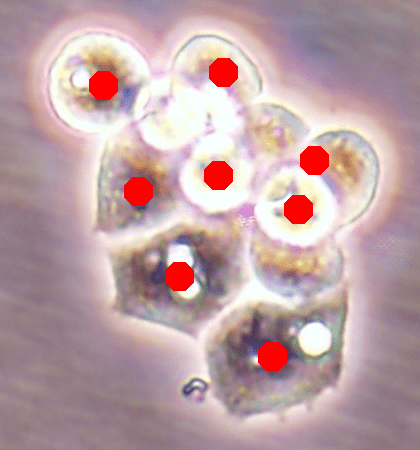} & 
\includegraphics[width =0.234\columnwidth]{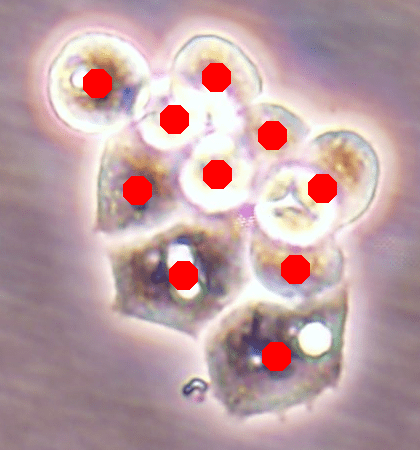} \\ 
\begin{sideways}\bf{~~~Independent}\end{sideways} &
\includegraphics[width =0.234\columnwidth]{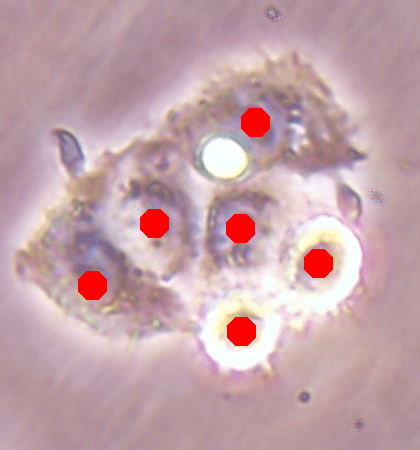} & 
\includegraphics[width =0.234\columnwidth]{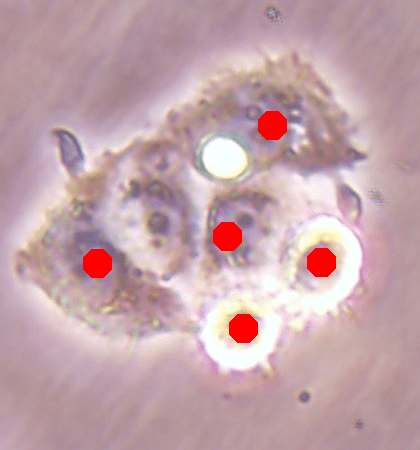} & 
\includegraphics[width =0.234\columnwidth]{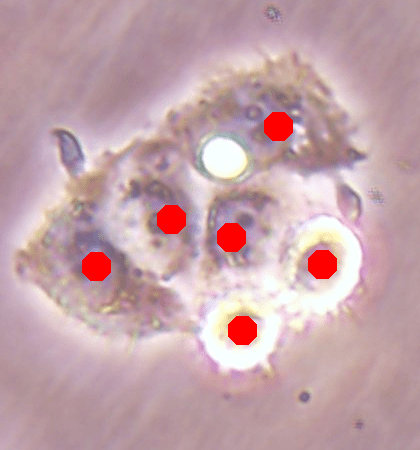} & 
\includegraphics[width =0.234\columnwidth]{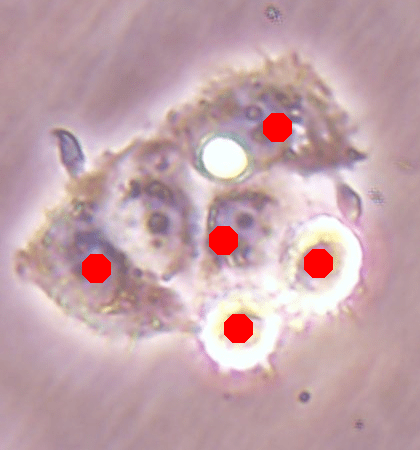} & 
\includegraphics[width =0.234\columnwidth]{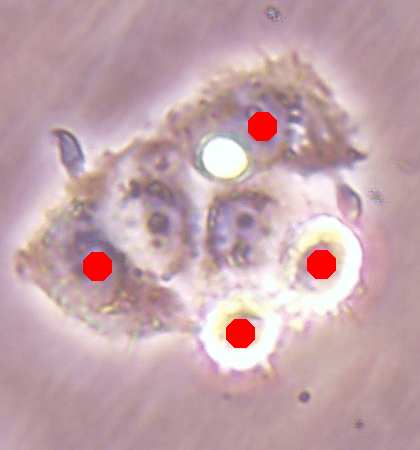} & 
\includegraphics[width =0.234\columnwidth]{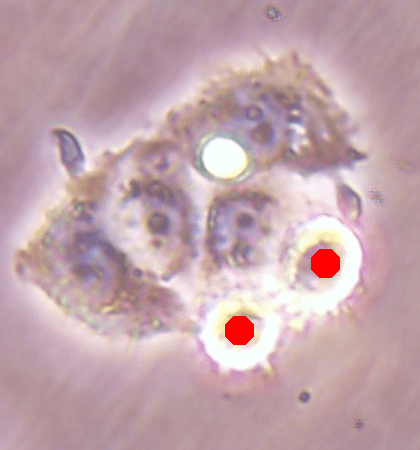} & 
\includegraphics[width =0.234\columnwidth]{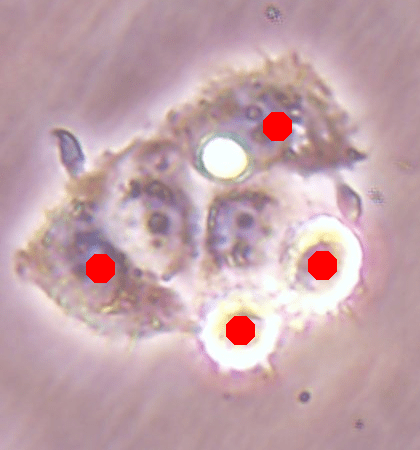} & 
\includegraphics[width =0.234\columnwidth]{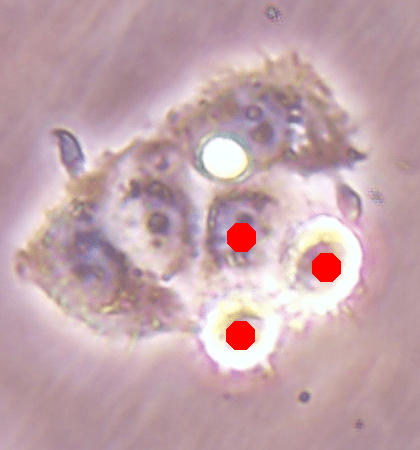} \\
\begin{sideways}\bf{~~~Independent}\end{sideways} &
\includegraphics[width =0.234\columnwidth]{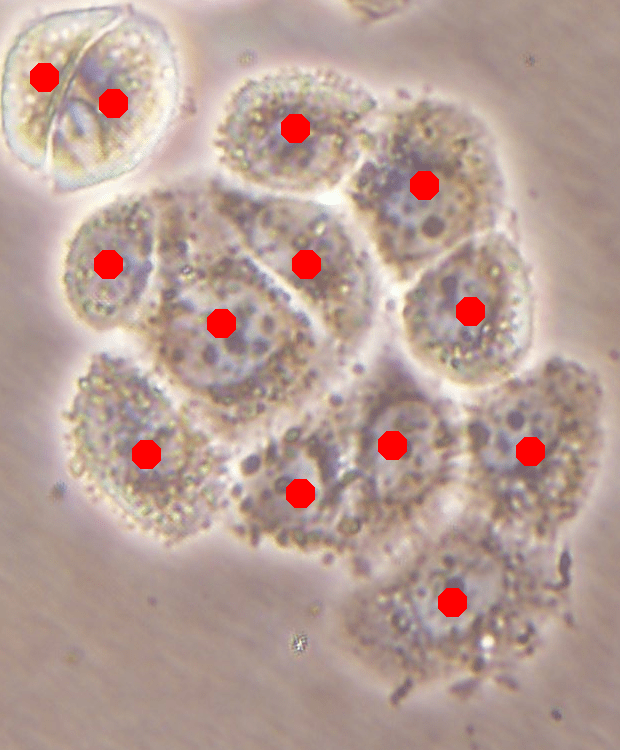}&
\includegraphics[width =0.234\columnwidth]{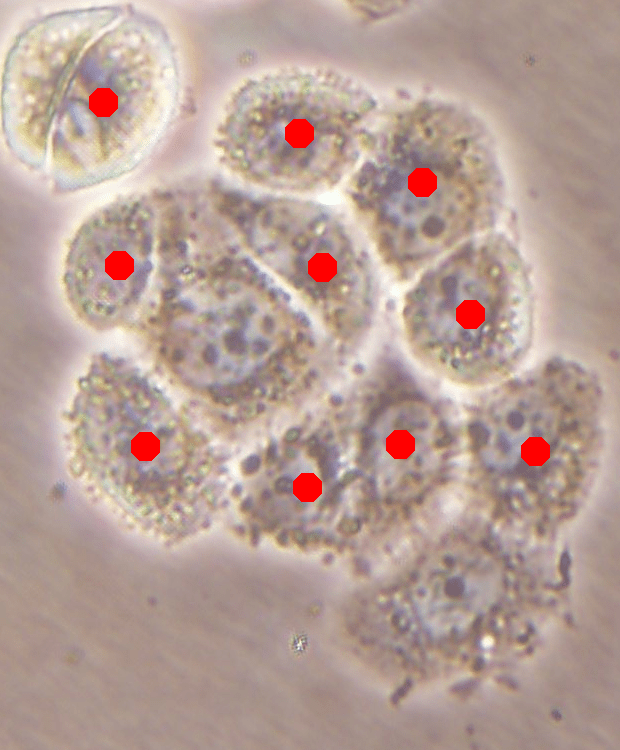}&
\includegraphics[width =0.234\columnwidth]{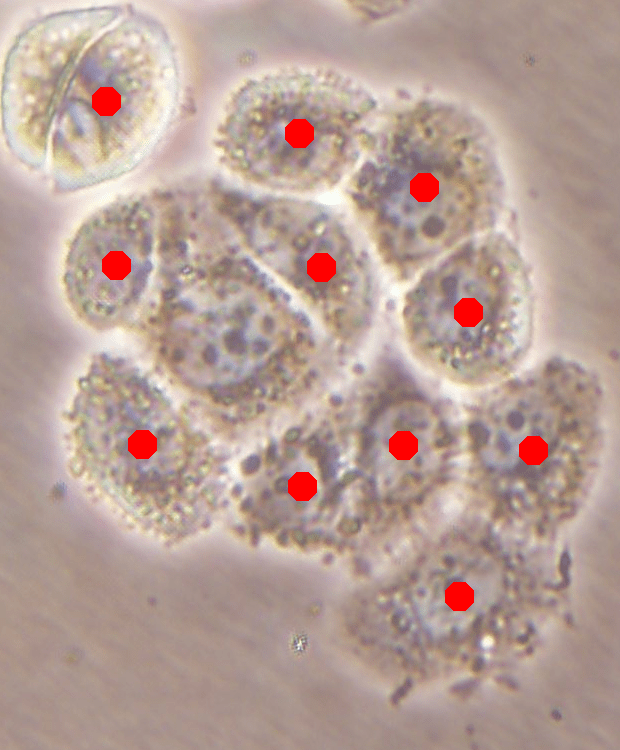}&
\includegraphics[width =0.234\columnwidth]{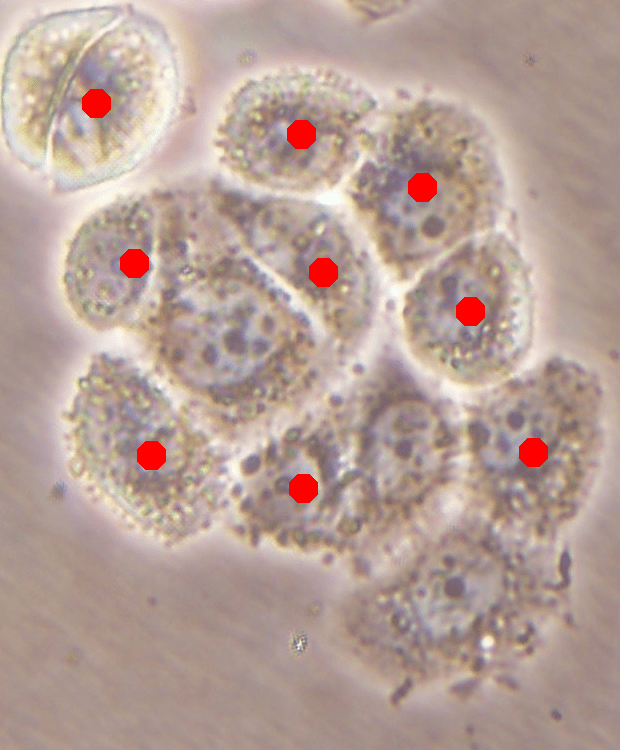}&
\includegraphics[width =0.234\columnwidth]{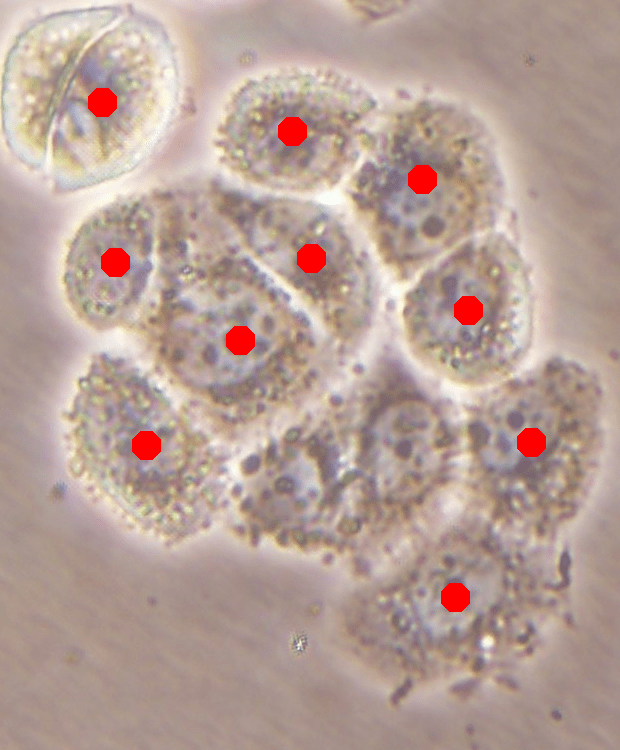}&
\includegraphics[width =0.234\columnwidth]{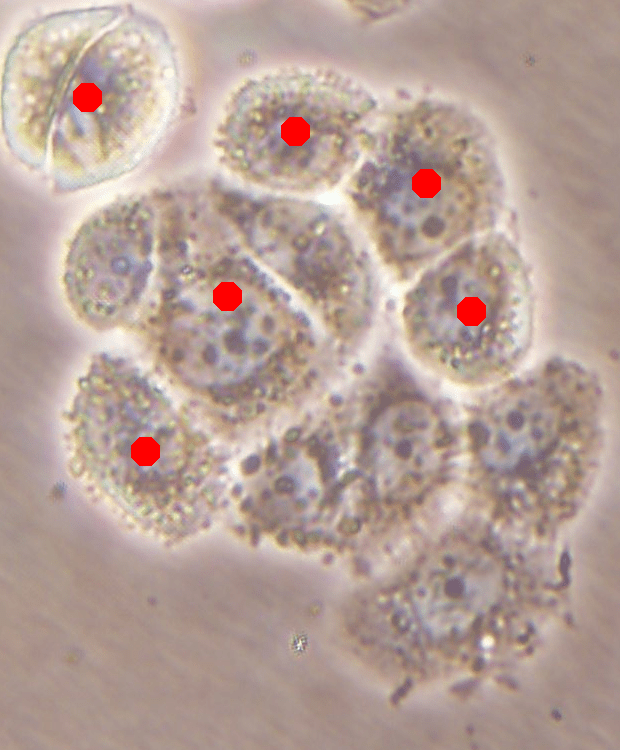}&
\includegraphics[width =0.234\columnwidth]{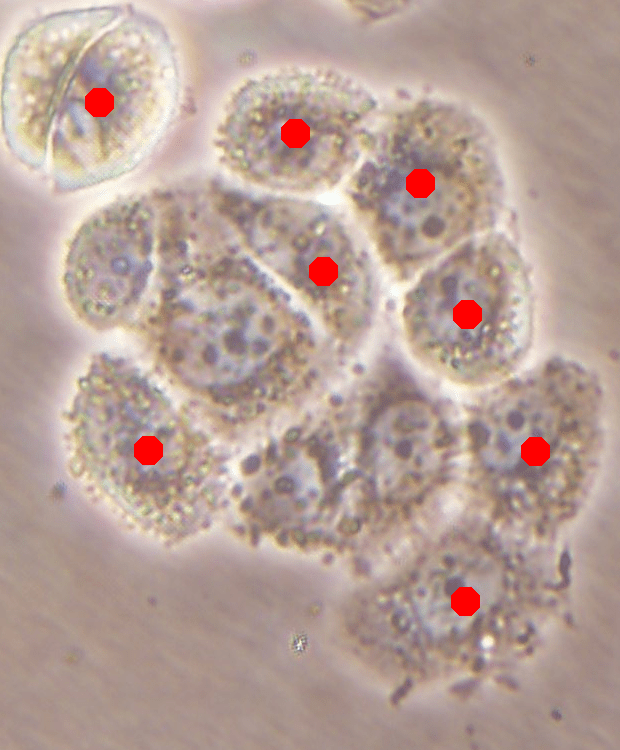}&
\includegraphics[width =0.234\columnwidth]{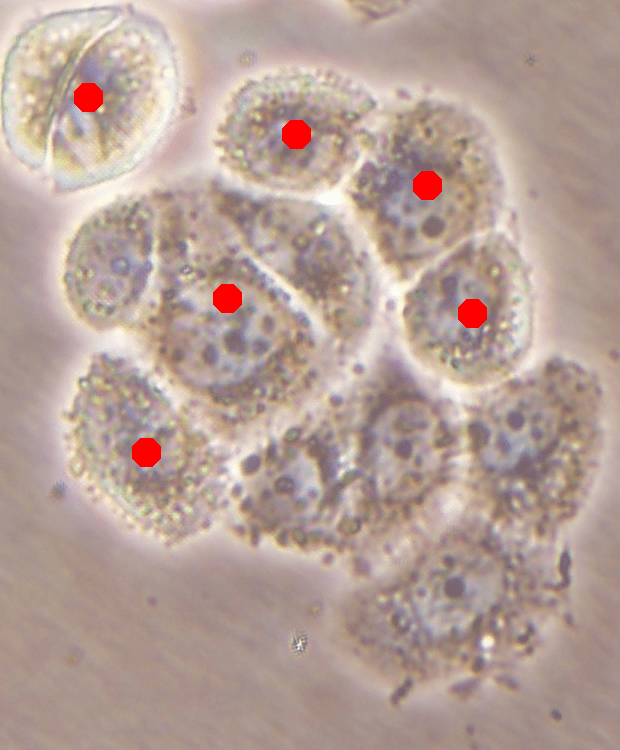} \\
 & (a) & (b) & (c) & (d) & (e) & (f) & (g) & (h)\\
\end{tabular}}
\caption{Visual results for illustrative subimages. (a) Annotated cells, (b)-(c) cells correctly identified by the proposed models, and (d)-(h) cells correctly identified by the comparison methods. In particular, cells correctly identified by (b) \textit{DeepDistance}, (c) the extended version of \textit{DeepDistance}, (d) SingleInner, (e) SingleOuter, (f) SingleClassification, (g) CascadedClassificationInner, and (h) MultiClassificationBoundary. Note that this figure shows only the cells correctly identified by the algorithms. It does not show any incorrectly located cell, which does not one-to-one match with any annotated marker. Also note that the first two subimages are taken from the dependent test samples and the last two from the independent test samples, which belong to the cell line that was not used in any part of the training.}
\label{fig:results}
\end{figure*}

\subsection{Results}

The quantitative results obtained by the proposed multi-task \textit{DeepDistance} models are given in Table~\ref{table:results}. This table reveals that our models lead to accurate cell detection results on both dependent and independent test samples. It also shows that extending the model by including the additional task of cell pixel classification further improves the results. Additionally, the visual results obtained on exemplary subimages are presented in Fig.~\ref{fig:results}. Note that this figure shows only the cells correctly identified by the models; it does not show any incorrectly located cell, which does not one-to-one match with any annotated marker. 

In order to understand the effectiveness of the \textit{DeepDistance} model, which uses a multi-task regression framework, we compare it with three deep learning based methods that use a single-task framework. These methods are designed to separately learn the tasks used by the proposed \textit{DeepDistance} models. In particular, they learn only an inner distance map (\textit{SingleInner}), only a normalized outer distance map (\textit{SingleOuter}), and only a classification map of cell pixels (\textit{SingleClassification}) from the RGB image, respectively. For learning their single-tasks, all these methods use an FCN with a single encoder path, similar to our models, but also only a single decoder path, as opposed to ours. The convolution and pooling/upsampling layers of this single encoder and single decoder path are the same with those specified in Fig.~\ref{fig:fcn-architecture}. They also end-to-end train their FCNs and the training setups are the same with ours. Obviously, since they have only one task to be learned, none of these methods take advantage of learning the shared feature representations. After learning their FCNs, these methods take the same cell detection steps of our model. These steps include estimating a map by the FCN, suppressing its noise by the h-maxima transform, and finding regional maxima on the resulting map. Here the \textit{SingleInner} and \textit{SingleOuter} methods use their estimated distance maps and the \textit{SingleClassification} method uses the estimated posterior map of the cell pixel class. The quantitative results of these methods are given in Table~\ref{table:results} and their visual results on exemplary subimages are presented in Fig.~\ref{fig:results}. These results reveal that concurrent learning of multiple tasks improves the results of single-task learning. It is worth to noting that this improvement is more evident for the independent test samples. 

Next, we compare \textit{DeepDistance} with two more methods that use more than one task in designing their models. The first one is the \textit{CascadedClassificationInner} method  that uses a cascaded architecture similar to the one proposed by~\cite{ram18}. This cascaded architecture is designed to sequentially learn a classification map of cell pixels from the RGB image and then to regress an inner distance map from the classification map. This method learns these maps in serial by using two serially cascaded FCNs that do not share any feature representation. More specifically, the first FCN takes the RGB image as the input and outputs the classification map whereas the second one takes the classification map as its input and outputs the inner distance map. Each of these serial FCNs has a single encoder and a single decoder path that contains the same convolution and pooling/upsampling layers specified in Fig.~\ref{fig:fcn-architecture}. Although these two FCNs are learned at the same time by backpropagating the error through the entire network in an end-to-end training fashion, they do not learn any shared feature representation. In other words, each FCN has its own encoder path. This method uses the same training setup with our model and takes the same cell detection steps. It detects the cells on the estimated inner distance map. That is, it first suppresses the noise in this estimated map with the h-maxima transform and then finds regional maxima on the noise suppressed map. The results given in Table~\ref{table:results} and Fig.~\ref{fig:results} demonstrate that the proposed \textit{DeepDistance} model, which learns multiple tasks in parallel with shared feature representations, gives more accurate results than the \textit{CascadedClassificationInner} method, which learns two tasks in serial without sharing any representation. This indicates the effectiveness of learning shared feature representations from multiple tasks, which is indeed known to be effective for many domains~\citep{caruana97}.

The last comparison is with the \textit{MultiClassificationBoundary} method that approaches cell detection as a multi-task classification problem. Similar to the model proposed by~\cite{chen17}, this method defines two classification tasks, where one is the task of cell pixel classification and the other is the task of cell boundary classification, and learns them in parallel by also using shared feature representations. For learning these tasks, this method uses an FCN whose architecture is the same with the one given in Fig.~\ref{fig:fcn-architecture}. This FCN is end-to-end trained also using the same training setup. To detect the cells in a given image, the \textit{MultiClassificationBoundary} method combines the two classification maps estimated by the FCN with a simple fusion technique that is also used by~\cite{chen17}. For that, the pixels estimated as boundary are subtracted from the estimated cell pixel classification map and the connected components on the resulting map are identified as cells. Since this fusion technique gives small noisy components, which lower the detection accuracy, the components smaller than an area threshold are eliminated. The value of this threshold is also selected on training and validation cells. The results of this method are given in Table~\ref{table:results} and Fig.~\ref{fig:results}. The table shows that this method, which approaches cell detection as a multi-task classification problem, also gives accurate results for the dependent test samples. On the other hand, for the independent test samples, it yields significantly lower accuracies than the proposed \textit{DeepDistance} model, which approaches cell detection as a multi-task regression problem. Note that these independent test samples belong to the cell line, any samples of which were not employed in any part of the training process. This significant difference may be attributed to the following. The two regression tasks learned by our model may convey more complementary information than the two classification tasks. Concurrent learning of more complementary tasks typically better helps the model less overfit on the training samples, and as a result, gives a more generalizable model. This may be the reason of our models better performing on the independent test samples taken from an unseen cell line. 

\subsection{Parameter Analysis}
\begin{figure}
\centering
\small{
\begin{tabular}{cc}
\includegraphics[width=0.46\columnwidth]{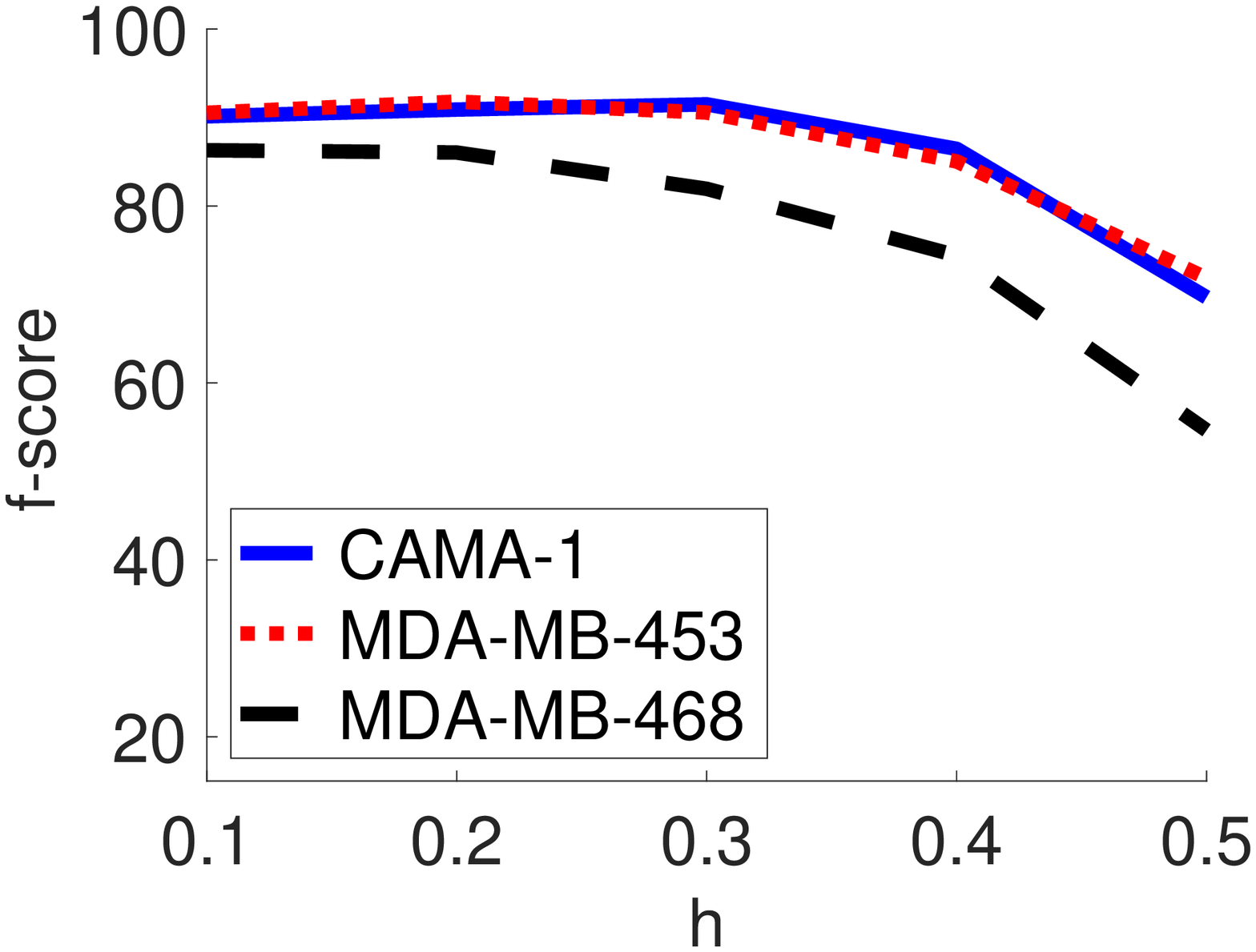} &
\includegraphics[width=0.46\columnwidth]{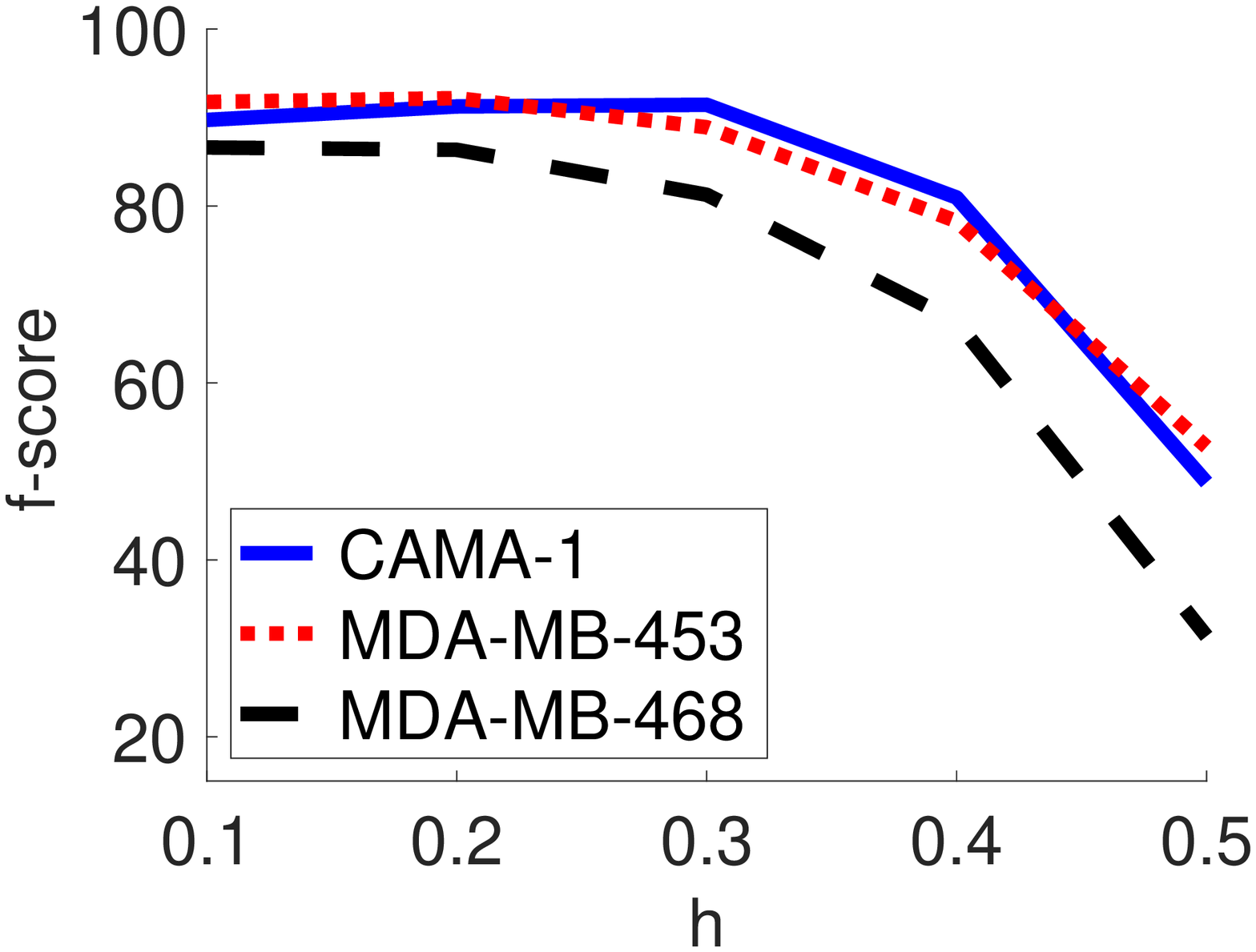} \\
(a) & (b)\\ 
\end{tabular}
}
\caption{Test set f-scores as a function of the $h$ parameter for (a) the \textit{DeepDistance}  model and (b) its extended version.}
\label{fig:param-analysis}
\end{figure}

The \textit{DeepDistance} model has one external parameter, the $h$ value used by the h-maxima transform to suppress noise in the estimated inner distance map. Small $h$ values do not sufficiently suppress the noise, resulting in false cell detections and oversegmentations. On the other hand, unnecessarily large values suppress too many pixels as the noise, causing not to identify many actual cells and leading to undersegmentations. Both of these cases decrease the performance. This is consistent with our experimental results shown in Figs.~\ref{fig:param-analysis}a and~\ref{fig:param-analysis}b, which depict the test set f-scores as a function of the $h$ value for the \textit{DeepDistance} model and its extended version, respectively. 

\subsection{Multi-task vs. Single-task Learning}

Since the main goal of this work is cell detection, our \textit{DeepDistance} models define the estimation of an inner distance map as the main task and find regional maxima on this estimated map to detect cells. The motivation behind these choices is the fact that the inner distance definition gives sharp increases at cell centers and the locations with these sharp increases can be detected by finding regional maxima. Hence, to obtain accurate detections, one should estimate an inner distance map with distinct differences between the cell centers and their surrounding pixels such that these centers can be identified as regional maxima. That is, one should estimate a map consisting of sharp enough bright regions close to the cell centers. To improve the performance of the task of this inner distance estimation, our models take advantage of multi-task learning approach. This approach helps the models become more robust to avoid overfitting a task, compared to the approach of learning the same task alone with a single-task model~\citep{caruana97}. To get more insight in this multi-task learning approach, this section visually analyzes the estimated maps of single-task and multi-task models. 

\begin{figure}
\centering
\small{
\begin{tabular}{@{~}c@{~}c@{~}c@{~}c@{~}}
\includegraphics[width=0.20\columnwidth]{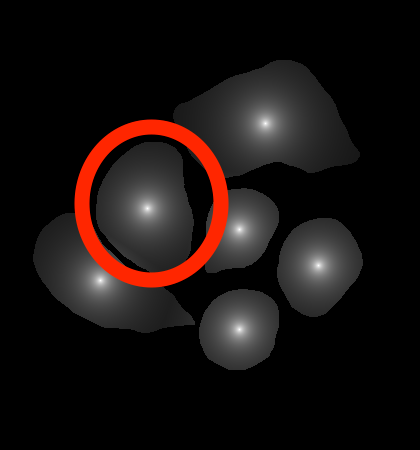} &
\includegraphics[width=0.20\columnwidth]{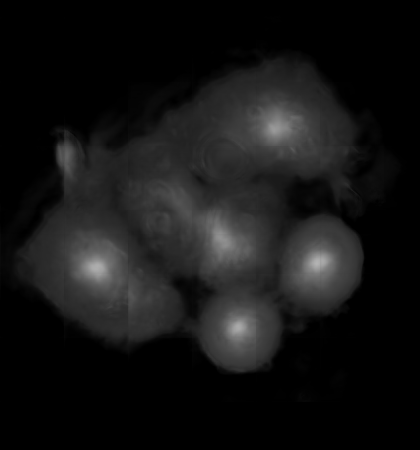} &
\includegraphics[width=0.20\columnwidth]{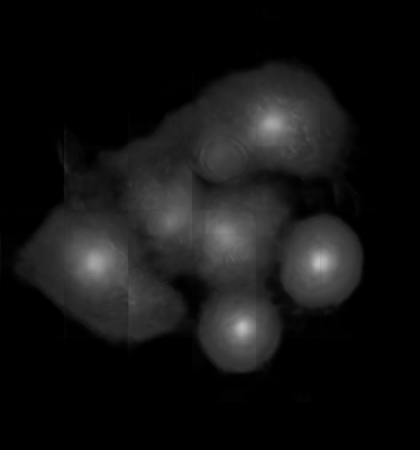} &
\includegraphics[width=0.20\columnwidth]{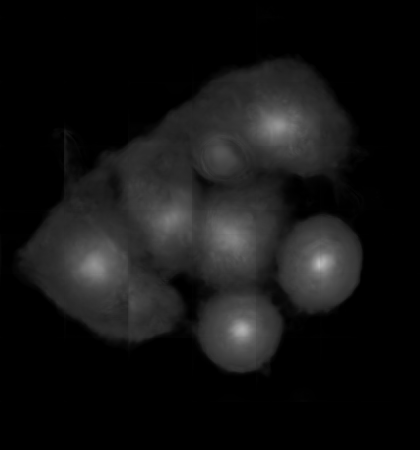} \\
\includegraphics[width=0.20\columnwidth]{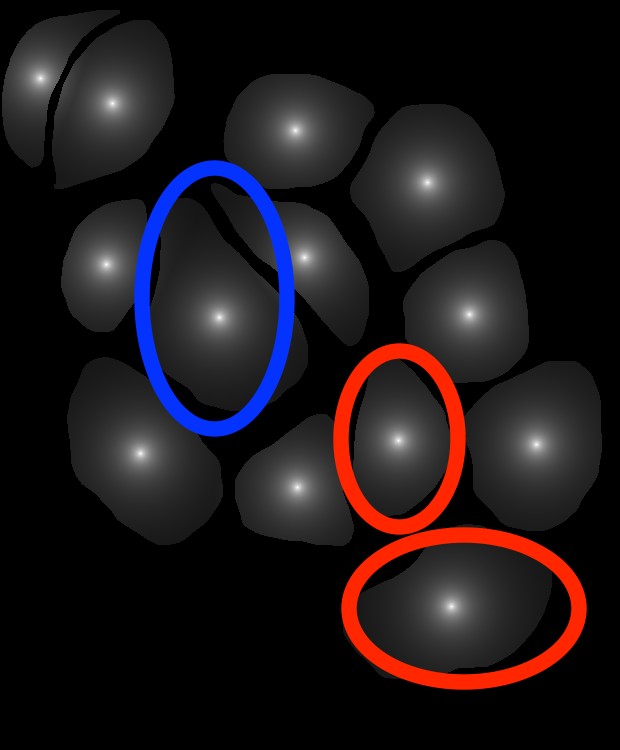} &
\includegraphics[width=0.20\columnwidth]{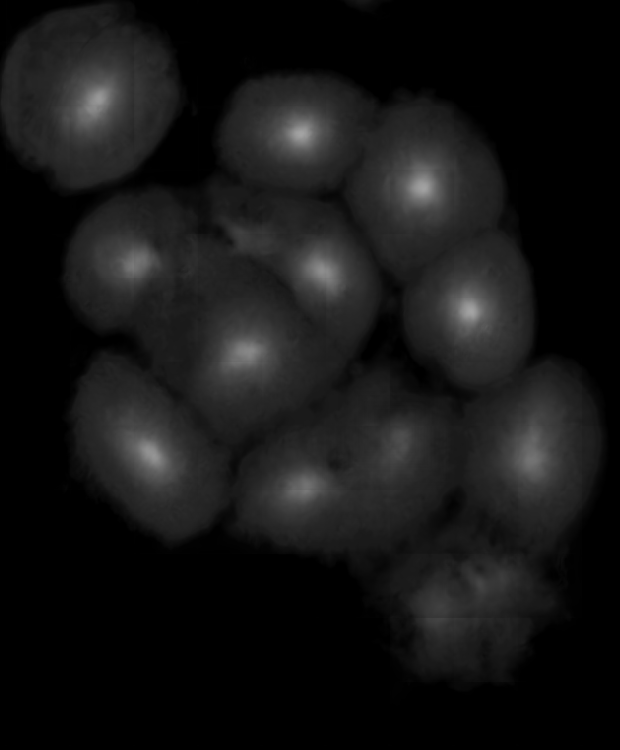} &
\includegraphics[width=0.20\columnwidth]{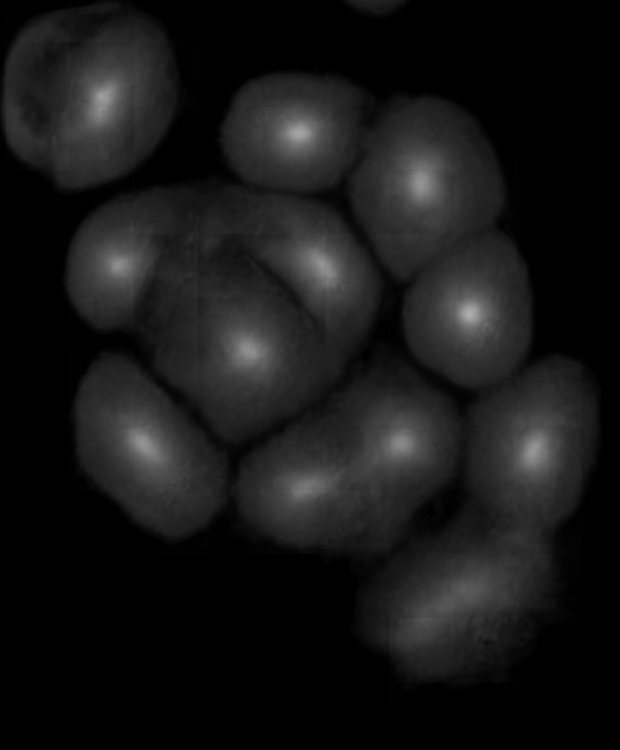} &
\includegraphics[width=0.20\columnwidth]{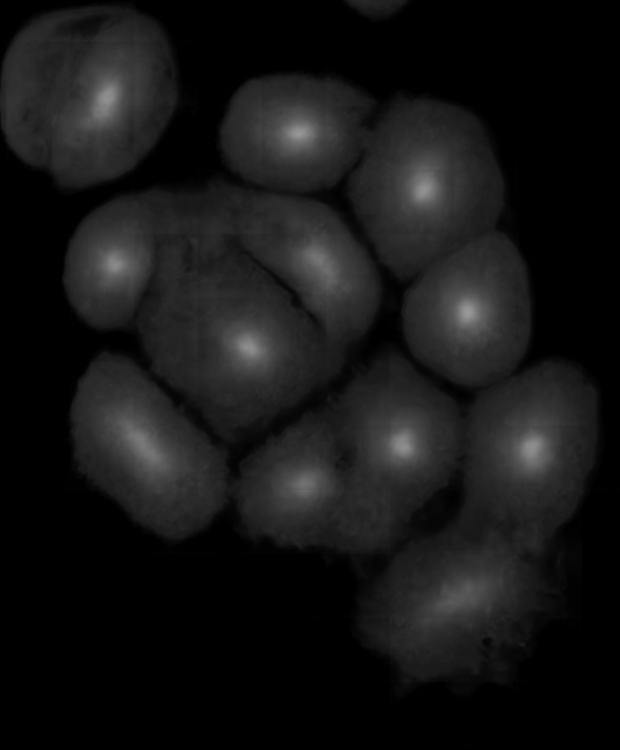} \\
(a) & (b) & (c) & (d)\\ 
\end{tabular}
}
\caption{(a) Maps of calculated inner distances when the ground truths are provided. Inner distance maps estimated by (b) \textit{SingleInner}, (c) \textit{DeepDistance}, and (d) the extended version of \textit{DeepDistance}.}
\label{fig:inner-single-vs-multi}
\end{figure}

For the independent test samples given in Fig.~\ref{fig:results}, Fig.~\ref{fig:inner-single-vs-multi}a shows the maps of the calculated inner distances when the ground truths are given. Figs.~\ref{fig:inner-single-vs-multi}b,~\ref{fig:inner-single-vs-multi}c, and~\ref{fig:inner-single-vs-multi}d illustrate the inner distance maps estimated by the \textit{SingleInner} method, the proposed \textit{DeepDistance} model, and its extended version, respectively. \textit{SingleInner} learns its map as a single-task whereas our models define auxiliary tasks and learn the inner distance map in parallel to these auxiliary tasks, forcing them to learn shared representations with a shared encoder path. The latter type of learning, which is an example of multi-task learning, is known to be effective for increasing the performance of individual tasks for many domains. We also observe this performance increase in the estimated maps given in Fig.~\ref{fig:inner-single-vs-multi}. \textit{SingleInner} cannot successfully detect the three cells shown inside red ellipses since it cannot produce sharp enough bright regions (with distinct enough estimated distances) for these cells. Although \textit{DeepDistance}, which uses one auxiliary task, leads to brighter regions for these cells, they are still not sharp enough for two of them to be identified as regional maxima. The extended version of \textit{DeepDistance}, which uses one more auxiliary task, does better job in inner distance estimations such that they have sharp enough bright regions for all of these three cells.

In this figure, it is worth to noting two points: First, all methods apply the h-maxima transform on their estimated maps beforehand to suppress noise, and hence, to prevent over-segmentations and false positives. If it was not applied, \textit{SingleInner} might give regional maxima for some of the three cells even though the distances estimated for their centers were not that distinct (bright). However, that case would also give many over-segmented cells and false positives. Second, none of the methods identify the cell shown inside a blue ellipse although their estimated distances yield bright regions for this cell. It is due to the evaluation method, which matches an annotated marker and a detected cell based on the distance between them since a test set image does not have boundary annotations but just a dot on each cell. In our experiments, a distance threshold is set to 30, considering image resolutions and the average cell size. This threshold may give a few incorrect matchings especially for larger cells, (e.g.,  the cell shown inside the blue ellipse). Increasing this threshold solves the problem for this particular cell, but this time, it will result in many incorrect matchings of detected cells with distant markers (or vice versa).

\begin{figure}
\centering
\small{
\begin{tabular}{@{~}c@{~}c@{~}c@{~}c@{~}}
\includegraphics[width=0.20\columnwidth]{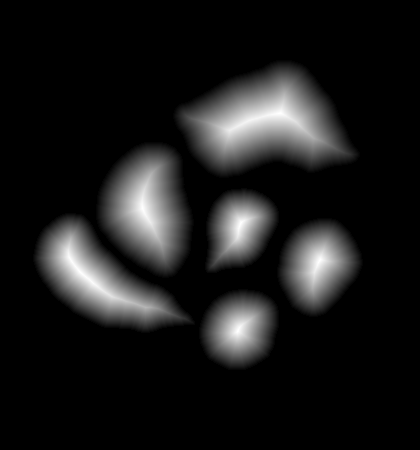} &
\includegraphics[width=0.20\columnwidth]{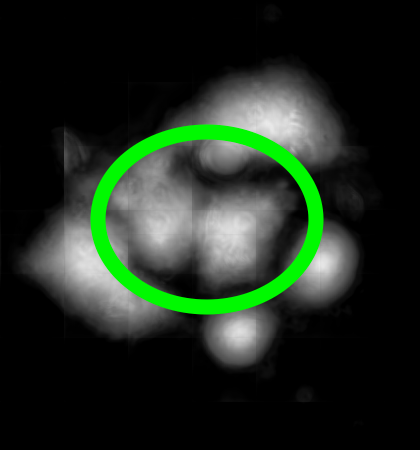} &
\includegraphics[width=0.20\columnwidth]{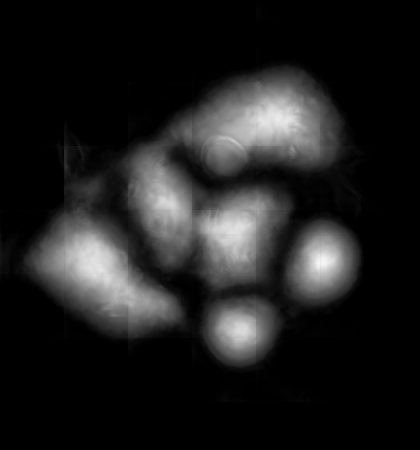} &
\includegraphics[width=0.20\columnwidth]{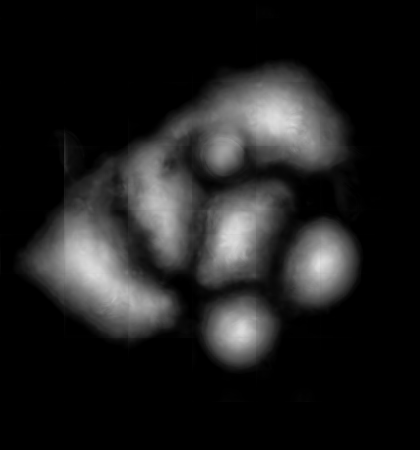} \\
\includegraphics[width=0.20\columnwidth]{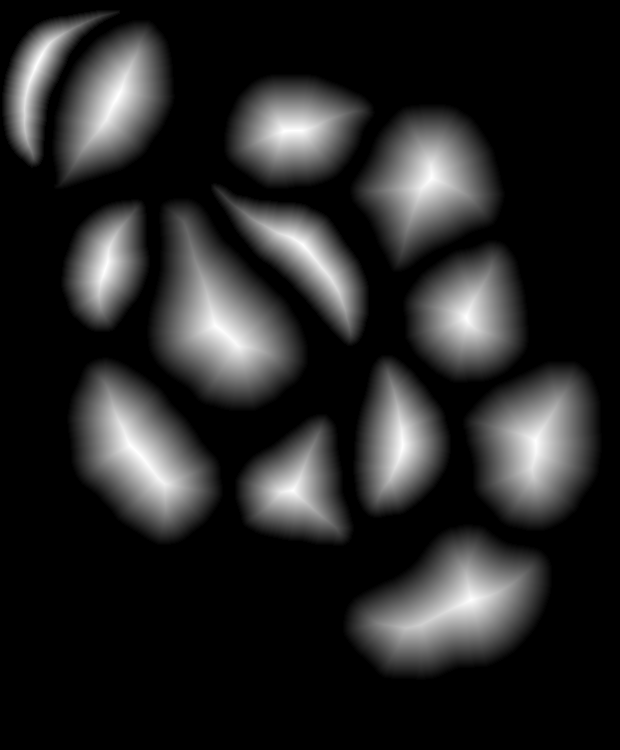} &
\includegraphics[width=0.20\columnwidth]{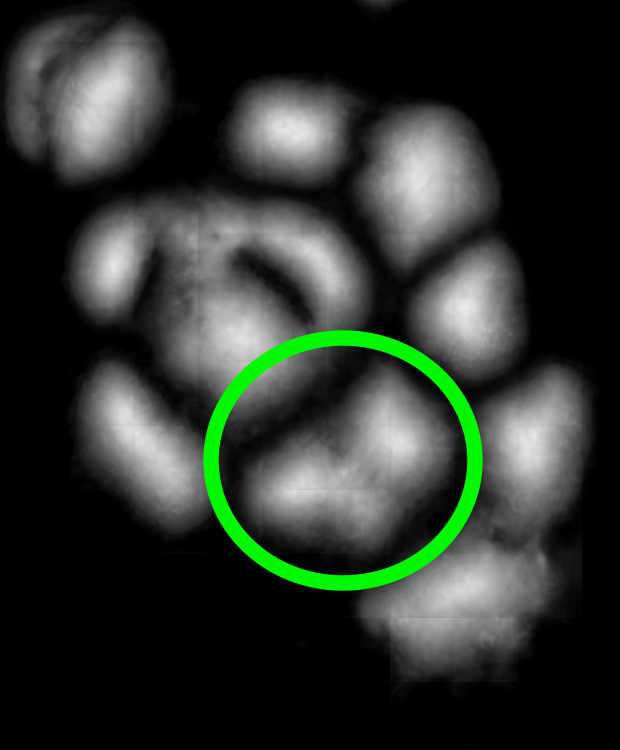} &
\includegraphics[width=0.20\columnwidth]{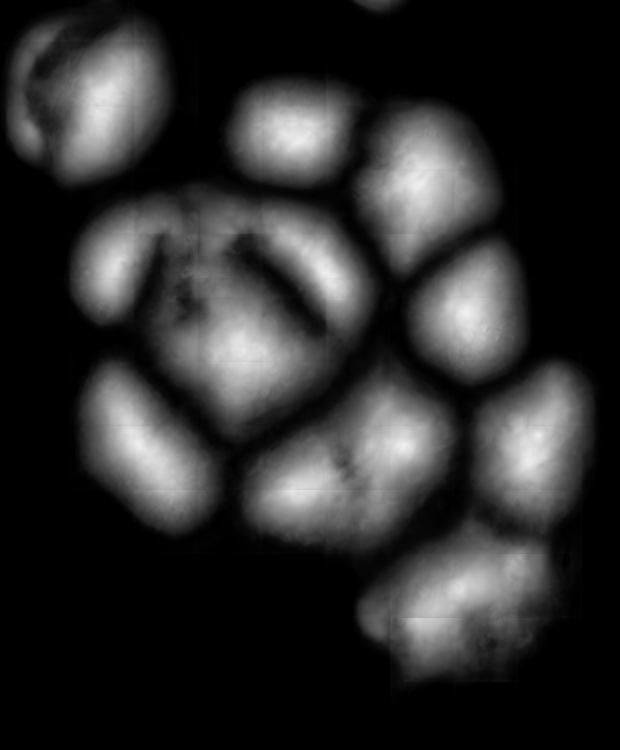} &
\includegraphics[width=0.20\columnwidth]{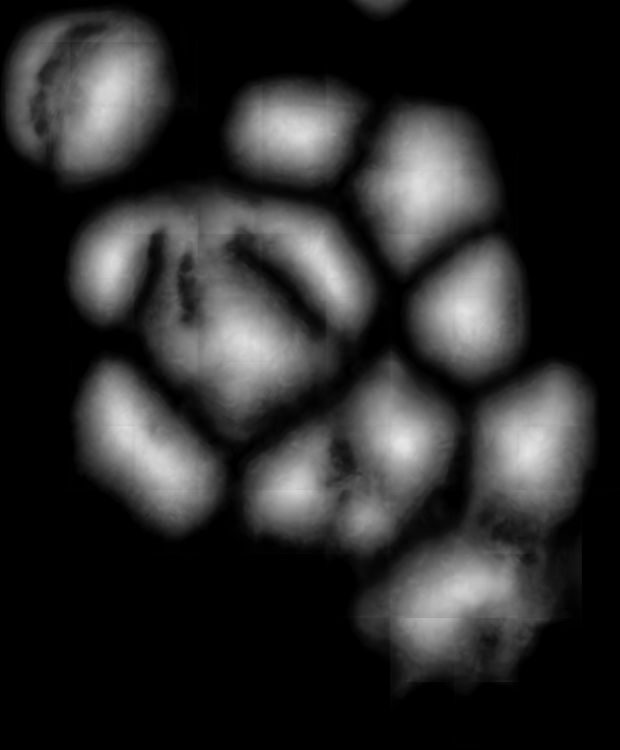} \\
(a) & (b) & (c) & (d)\\ 
\end{tabular}
}
\caption{(a) Maps of calculated outer distances when the ground truths are provided. Note that test set images do not have boundary annotations; we draw the boundaries for these two samples for illustration purposes. Outer distance maps estimated by (b) \textit{SingleOuter}, (c) \textit{DeepDistance}, and (d) the extended version of \textit{DeepDistance}.}
\label{fig:outer-single-vs-multi}
\end{figure}

Likewise, Fig.~\ref{fig:outer-single-vs-multi}a shows the maps of calculated outer distances when the ground truths are given. Figs.~\ref{fig:outer-single-vs-multi}b,~\ref{fig:outer-single-vs-multi}c, and~\ref{fig:outer-single-vs-multi}d show the outer distance maps estimated by \textit{SingleOuter}, \textit{DeepDistance}, and its extended version, respectively. It is observed that a single-task \textit{SingleOuter} method is less accurate in estimating outer distances especially for pixels close to cell boundaries. Due to this incorrect estimation, it locates only a single cell for each of the cell pairs shown inside green ellipses, resulting in under-segmentations for these cell pairs. Our multi-task \textit{DeepDistance} models yield better estimations for these boundary pixels. However, it is important to note that our models do not use the estimated outer distances in a detection algorithm but define this estimation as an auxiliary task. Particularly, this distance is defined to represent a different aspect of the problem and its estimation is considered as complementary to the main task. Concurrent learning of two related tasks with a multi-task model, which uses shared feature representations, better helps avoid local optima. In other words, when two related tasks share the same representations (with a shared encoder path), it is more difficult to finetune these representations for only one of these tasks. This is effective to obtain better learning performances for individual tasks, as also shown in Figs.~\ref{fig:inner-single-vs-multi} and~\ref{fig:outer-single-vs-multi}.

\subsection{Extending DeepDistance for Cell Segmentation}
\label{sec:segmentation}

The \textit{DeepDistance} models are designed to identify individual cells by locating their centers without delineating their boundaries. This goal is to address the problem of identifying cells for the purpose of counting (or tracking), which is a very common practice for cell culture research. For instance, cell counting can be used to determine the number of cells in a tissue culture in-real-time within predefined intervals for examining the cell growth under the effects of a cytotoxic treatment. This section discusses how the outputs of \textit{DeepDistance} can be used for cell segmentation. To this end, it presents a simple marker-controlled region growing algorithm that works on the outputs of the extended \textit{DeepDistance} model.

\begin{figure}[!t]
\centering
\footnotesize{
\begin{tabular}{@{~}c@{~}c@{~}c@{~}c@{~}}
\includegraphics[height =0.262\columnwidth]{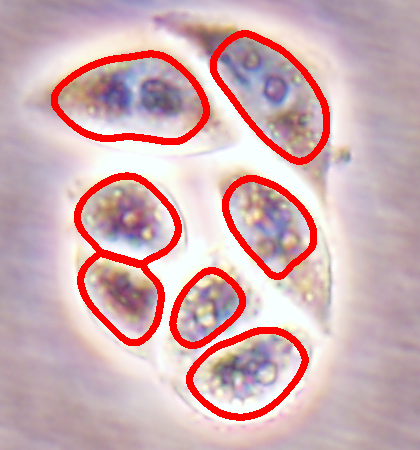} & 
\includegraphics[height =0.262\columnwidth]{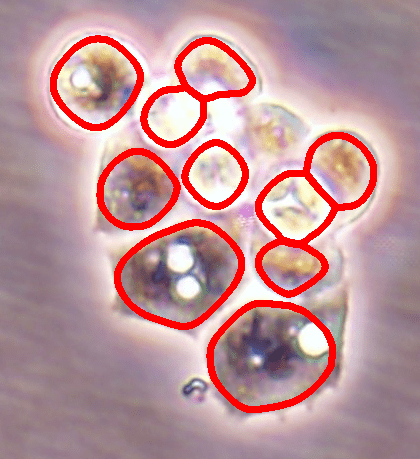} & 
\includegraphics[height =0.262\columnwidth]{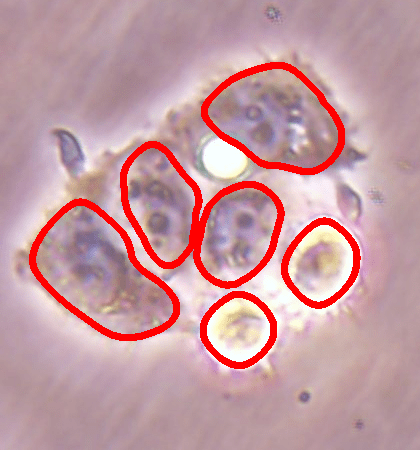} & 
\includegraphics[height =0.262\columnwidth]{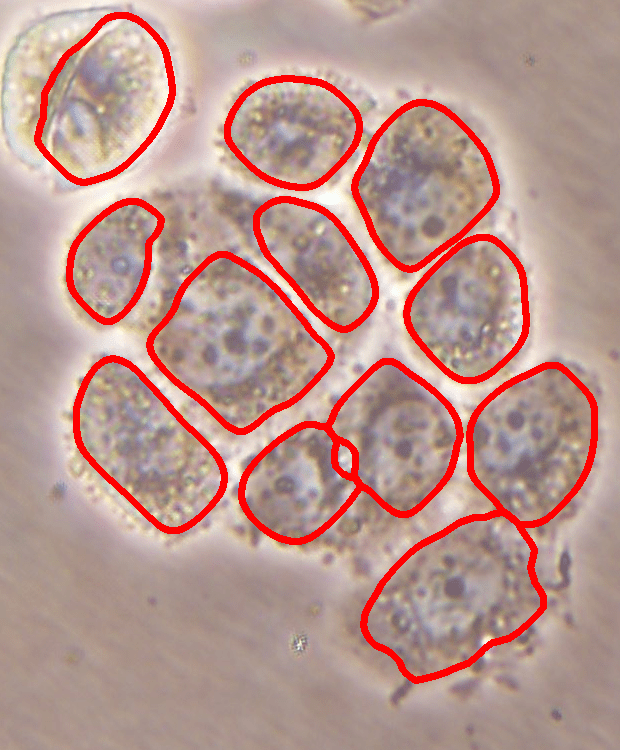} \\
\end{tabular}}
\caption{Visual segmentation results for illustrative subimages, which detection results are shown in Fig.~\ref{fig:results}.}
\label{fig:seg-results}
\end{figure}

This algorithm has two steps: marker identification and marker growing. Let ${\cal M}_I$ and ${\cal M}_O$ be the estimated maps of the inner and normalized outer distances, respectively, and ${\cal M}_C$ be the estimated classification map. The marker identification step takes the cell locations identified on ${\cal M}_I$ by \textit{DeepDistance} and slightly widens them on ${\cal M}_O$ to obtain better-shaped markers. Since the aim is to obtain better shapes for the markers but not to get their exact boundaries, which will be done by the next step, the markers are widened onto only the pixels close to cell centers but not close to boundaries, by considering only the pixels whose estimated values in ${\cal M}_O > 0.5$. At the end, all widened markers smaller than an area threshold of 50 are eliminated. The next step iteratively grows the remaining markers onto foreground pixels in ${\cal M}_C$ with respect to distances in ${\cal M}_O$. This growing is marker-based such that each iteration selects the ``best'' marker and grows it onto the foreground pixels that are currently adjacent to this selected marker. To do so, for each marker, it calculates the average ${\cal M}_O$ over its adjacent foreground pixels and selects the marker with the highest average. This selection does not consider a marker if its adjacent foreground pixels are less than the half of all pixels that are adjacent to this marker. The growing process stops when there exists no marker to be selected. At the end of the marker growing step, boundaries of the grown markers are smoothed applying a majority filter with a size of 25. The visual results obtained on the exemplary subimages are presented in Fig.~\ref{fig:seg-results}.

This section presents a simple algorithm to discuss the possibility of using the estimated maps for segmentation. However, it is worth to noting that this work designs its multi-task network and the tasks used in this network for the purpose of facilitating the cell detection problem (not the cell segmentation problem). Although the estimated normalized outer distance and classification maps provide some information about the cell boundaries, the primary goal of these maps is to define auxiliary tasks, concurrent learning of which is expected to improve the performance of the main task in the context of multi-task learning~\citep{caruana97}. On the other hand, for more accurate cell segmentation, one needs to design a network that includes additional tasks specifically focused on learning the cell boundaries (e.g., defining the estimation of the cell boundaries as the main task) and to use the estimated boundaries in the region growing process too. Designing such multi-task networks, developing more sophisticated cell segmentation algorithms that work on the outputs of these networks, and conducting systematic experiments to obtain their quantitative results are considered as future work.

\section{Conclusion}

This paper presents the \textit{DeepDistance} model, which designs a multi-task regression framework for detecting individual cells in microscopy images, and experimentally demonstrates the successful use of this model on 5192 cells of three different cell lines. For the cell detection problem, this is the first proposal of a multi-task regression model that learns multiple regression tasks in parallel by using shared feature representations.

The \textit{DeepDistance} model designs this regression framework to concurrently learn two distance metrics for image pixels in the context of multi-task learning. To this end, it defines the \textit{normalized outer distance} metric to represent a different aspect of the problem and proposes to learn it in parallel to the primary inner distance metric, which is defined in regard to the main cell detection task, for the purpose of increasing the generalization ability of this main task. For this concurrent learning, the \textit{DeepDistance} model constructs a fully convolutional network (FCN) with a shared encoder path, which forces the two tasks to learn shared feature representations at various abstraction levels. Such shared representation learning on multiple tasks is indeed known to be more effective to avoid each task to overfit, and as a result, to obtain more generalized models. Our experiments on three different cell lines also reveal that this multi-task learning together with formulating cell detection as a regression problem lead to accurate results, improving the results of the single-task frameworks as well as the previous deep learning approaches.

This work mainly focuses on the cell detection problem, which corresponds to identifying cell locations in an unannotated image. It does not focus on delineating the precise boundaries of the cells. Designing multi-task networks that include additional tasks specifically focused on cell segmentation and developing algorithms that work on the outputs of these networks are considered as one of the future research directions of this study. We believe that the cell detection method proposed by this study is not limited to static cell images, but it can easily be adapted and used for real-time live images throughout a time dependent experiment under inverted microscopy. The investigation of the latter use is considered as another future research direction of this study.

\section*{Acknowledgments}
This work was supported by the Turkish Academy of Sciences under the Distinguished Young Scientist Award Program (T{\"U}BA GEB\.{I}P). 

\bibliography{bib}

\begin{thebibliography}{31}
\expandafter\ifx\csname natexlab\endcsname\relax\def\natexlab#1{#1}\fi
\providecommand{\url}[1]{\texttt{#1}}
\providecommand{\href}[2]{#2}
\providecommand{\path}[1]{#1}
\providecommand{\DOIprefix}{doi:}
\providecommand{\ArXivprefix}{arXiv:}
\providecommand{\URLprefix}{URL: }
\providecommand{\Pubmedprefix}{pmid:}
\providecommand{\doi}[1]{\href{http://dx.doi.org/#1}{\path{#1}}}
\providecommand{\Pubmed}[1]{\href{pmid:#1}{\path{#1}}}
\providecommand{\bibinfo}[2]{#2}
\ifx\xfnm\relax \def\xfnm[#1]{\unskip,\space#1}\fi

\bibitem[{Bai and Urtasun (2017)}]{bai17}
\bibinfo{author}{Bai, M.}, \bibinfo{author}{Urtasun, R.}, \bibinfo{year}{2017}.
\newblock \bibinfo{title}{Deep watershed transform for instance segmentation}, in: \bibinfo{booktitle}{Proc. of the IEEE Conf. on
  Computer Vision and Pattern Recognition}, pp. \bibinfo{pages}{5221--5229}.

\bibitem[{Caruana(1997)}]{caruana97}
\bibinfo{author}{Caruana, R.}, \bibinfo{year}{1997}.
\newblock \bibinfo{title}{Multitask learning}.
\newblock \bibinfo{journal}{Machine Learning} \bibinfo{volume}{28},
  \bibinfo{pages}{41--75}.
\bibitem[{Chang et~al.(2013)Chang, Han, Borowsky, Loss, Gray, Spellman and
  Parvin}]{chang13}
\bibinfo{author}{Chang, H.}, \bibinfo{author}{Han, J.},
  \bibinfo{author}{Borowsky, A.}, \bibinfo{author}{Loss, L.},
  \bibinfo{author}{Gray, J.W.}, \bibinfo{author}{Spellman, P.T.},
  \bibinfo{author}{Parvin, B.}, \bibinfo{year}{2013}.
\newblock \bibinfo{title}{Invariant delineation of nuclear architecture in
  glioblastoma multiforme for clinical and molecular association}.
\newblock \bibinfo{journal}{IEEE Transactions on Medical Imaging}
  \bibinfo{volume}{32}, \bibinfo{pages}{670--682}.
\bibitem[{Chen et~al.(2017)Chen, Qi, Yu, Dou, Qin and Heng}]{chen17}
\bibinfo{author}{Chen, H.}, \bibinfo{author}{Qi, X.}, \bibinfo{author}{Yu, L.},
  \bibinfo{author}{Dou, Q.}, \bibinfo{author}{Qin, J.}, \bibinfo{author}{Heng,
  P.A.}, \bibinfo{year}{2017}.
\newblock \bibinfo{title}{DCAN: Deep contour-aware networks for object instance
  segmentation from histology images}.
\newblock \bibinfo{journal}{Medical Image Analysis} \bibinfo{volume}{36},
  \bibinfo{pages}{135--146}.
\bibitem[{Chen et~al.(2016)Chen, Wang and Heng}]{chen16}
\bibinfo{author}{Chen, H.}, \bibinfo{author}{Wang, X.}, \bibinfo{author}{Heng,
  P.A.}, \bibinfo{year}{2016}.
\newblock \bibinfo{title}{Automated mitosis detection with deep regression
  networks}, in: \bibinfo{booktitle}{IEEE 13th
  Int. Symp. on Biomedical Imaging}, pp.
  \bibinfo{pages}{1204--1207}.
\bibitem[{Ciresan et~al.(2013)Ciresan, Giusti, Gambardella and
  Schmidhuber}]{ciresan13}
\bibinfo{author}{Ciresan, D.C.}, \bibinfo{author}{Giusti, A.},
  \bibinfo{author}{Gambardella, L.M.}, \bibinfo{author}{Schmidhuber, J.},
  \bibinfo{year}{2013}.
\newblock \bibinfo{title}{Mitosis detection in breast cancer histology images
  with deep neural networks}, in: \bibinfo{booktitle}{Int. Conf.
  on Medical Image Computing and Computer-Assisted Intervention},
  \bibinfo{organization}{Springer}, pp. \bibinfo{pages}{411--418}.
\bibitem[{Dima et~al.(2011)Dima, Elliott, Filliben, Halter, Peskin, Bernal,
  Kociolek, Brady, Tang and Plant}]{dima11}
\bibinfo{author}{Dima, A.A.}, \bibinfo{author}{Elliott, J.T.},
  \bibinfo{author}{Filliben, J.J.}, \bibinfo{author}{Halter, M.},
  \bibinfo{author}{Peskin, A.}, \bibinfo{author}{Bernal, J.},
  \bibinfo{author}{Kociolek, M.}, \bibinfo{author}{Brady, M.C.},
  \bibinfo{author}{Tang, H.C.}, \bibinfo{author}{Plant, A.L.},
  \bibinfo{year}{2011}.
\newblock \bibinfo{title}{Comparison of segmentation algorithms for
  fluorescence microscopy images of cells}.
\newblock \bibinfo{journal}{Cytometry Part A} \bibinfo{volume}{79},
  \bibinfo{pages}{545--559}.
\bibitem[{Dong et~al.(2015)Dong, Shao, Da~Costa, Bandmann and Frangi}]{dong15}
\bibinfo{author}{Dong, B.}, \bibinfo{author}{Shao, L.},
  \bibinfo{author}{Da~Costa, M.}, \bibinfo{author}{Bandmann, O.},
  \bibinfo{author}{Frangi, A.F.}, \bibinfo{year}{2015}.
\newblock \bibinfo{title}{Deep learning for automatic cell detection in
  wide-field microscopy zebrafish images}, in: \bibinfo{booktitle}{IEEE 12th
  Int. Symp. on Biomedical Imaging}, pp. \bibinfo{pages}{772--776}.

\bibitem[{Gao et~al.(2014)}]{gao14}
\bibinfo{author}{Gao, Y.}, \bibinfo{author}{Wang, L.}, \bibinfo{author}{Shao,
  Y.}, \bibinfo{author}{Shen, D.}, \bibinfo{year}{2014}.
\newblock \bibinfo{title}{Learning distance transform for boundary detection and deformable segmentation in CT prostate images}.
\newblock \bibinfo{journal}{Int. Workshop on Machine Learning in Medical Imaging} \bibinfo{volume}{8679}, \bibinfo{pages}{93–100}.

\bibitem[{Genctav et~al.(2012)Genctav, Aksoy and
  Onder}]{genctav12}
\bibinfo{author}{Genctav, A.}, \bibinfo{author}{Aksoy, S.},
  \bibinfo{author}{Onder, S.}, \bibinfo{year}{2012}.
\newblock \bibinfo{title}{Unsupervised segmentation and classification of
  cervical cell images}.
\newblock \bibinfo{journal}{Pattern Recognition} \bibinfo{volume}{45},
  \bibinfo{pages}{4151--4168}.
\bibitem[{Jung and Kim(2010)}]{jung10}
\bibinfo{author}{Jung, C.}, \bibinfo{author}{Kim, C.}, \bibinfo{year}{2010}.
\newblock \bibinfo{title}{Segmenting clustered nuclei using h-minima
  transform-based marker extraction and contour parameterization}.
\newblock \bibinfo{journal}{IEEE Transactions on Biomedical Engineering}
  \bibinfo{volume}{57}, \bibinfo{pages}{2600--2604}.
\bibitem[{Kainz et~al.(2015)Kainz, Urschler, Schulter, Wohlhart and
  Lepetit}]{kainz15}
\bibinfo{author}{Kainz, P.}, \bibinfo{author}{Urschler, M.},
  \bibinfo{author}{Schulter, S.}, \bibinfo{author}{Wohlhart, P.},
  \bibinfo{author}{Lepetit, V.}, \bibinfo{year}{2015}.
\newblock \bibinfo{title}{You should use regression to detect cells}, in:
  \bibinfo{booktitle}{Int. Conf. on Medical Image Computing and
  Computer-Assisted Intervention}, \bibinfo{organization}{Springer}, pp.
  \bibinfo{pages}{276--283}.


\bibitem[{Kechyn (2018) Kechyn}]{kechyn18}
\bibinfo{author}{Kechyn, G.}, \bibinfo{year}{2018}.
\newblock \bibinfo{title}{Instance segmentation: automatic nucleus detection}, Towards Data Science, 
https://towardsdatascience.com/instance-segmentation-automatic-nucleus-detection-a169b3a99477.
%
\bibitem[{Koyuncu et~al.(2018)Koyuncu, , Cetin-Atalay and
  Gunduz-Demir}]{koyuncu18}
\bibinfo{author}{Koyuncu, C.F.}, \bibinfo{author}{Cetin-Atalay, R.},
  \bibinfo{author}{Gunduz-Demir, C.}, \bibinfo{year}{2018}.
\newblock \bibinfo{title}{Object-oriented segmentation of cell nuclei in
  fluorescence microscopy images}.
\newblock \bibinfo{journal}{Cytometry Part A}.
\bibitem[{Koyuncu et~al.(2016)Koyuncu, Akhan, Ersahin, Cetin-Atalay and
  Gunduz-Demir}]{koyuncu16}
\bibinfo{author}{Koyuncu, C.F.}, \bibinfo{author}{Akhan, E.},
  \bibinfo{author}{Ersahin, T.}, \bibinfo{author}{Cetin-Atalay, R.},
  \bibinfo{author}{Gunduz-Demir, C.}, \bibinfo{year}{2016}.
\newblock \bibinfo{title}{Iterative h-minima-based marker-controlled watershed
  for cell nucleus segmentation}.
\newblock \bibinfo{journal}{Cytometry Part A} \bibinfo{volume}{89},
  \bibinfo{pages}{338--349}.
\bibitem[{Litjens et~al.(2017)Litjens, Kooi, Bejnordi, Setio, Ciompi,
  Ghafoorian, van~der Laak, Van~Ginneken and Sanchez}]{litjens17}
\bibinfo{author}{Litjens, G.}, \bibinfo{author}{Kooi, T.},
  \bibinfo{author}{Bejnordi, B.E.}, \bibinfo{author}{Setio, A.A.A.},
  \bibinfo{author}{Ciompi, F.}, \bibinfo{author}{Ghafoorian, M.},
  \bibinfo{author}{van~der Laak, J.A.}, \bibinfo{author}{Van~Ginneken, B.},
  \bibinfo{author}{Sanchez, C.I.}, \bibinfo{year}{2017}.
\newblock \bibinfo{title}{A survey on deep learning in medical image analysis}.
\newblock \bibinfo{journal}{Medical Image Analysis} \bibinfo{volume}{42},
  \bibinfo{pages}{60--88}.
\bibitem[{Long et~al.(2015)Long, Shelhamer and Darrell}]{long15}
\bibinfo{author}{Long, J.}, \bibinfo{author}{Shelhamer, E.},
  \bibinfo{author}{Darrell, T.}, \bibinfo{year}{2015}.
\newblock \bibinfo{title}{Fully convolutional networks for semantic
  segmentation}, in: \bibinfo{booktitle}{Proc. of the IEEE Conf. on
  Computer Vision and Pattern Recognition}, pp. \bibinfo{pages}{3431--3440}.
\bibitem[{Ram et~al.(2018)Ram, Nguyen, Limesand and Sabuncu}]{ram18}
\bibinfo{author}{Ram, S.}, \bibinfo{author}{Nguyen, V.T.},
  \bibinfo{author}{Limesand, K.H.}, \bibinfo{author}{Sabuncu, M.R.},
  \bibinfo{year}{2018}.
\newblock \bibinfo{title}{Joint cell nuclei detection and segmentation in
  microscopy images using 3D convolutional networks}.
\newblock \bibinfo{journal}{arXiv preprint arXiv:1805.02850} .
\bibitem[{Ronneberger et~al.(2015)Ronneberger, Fischer and
  Brox}]{ronneberger15}
\bibinfo{author}{Ronneberger, O.}, \bibinfo{author}{Fischer, P.},
  \bibinfo{author}{Brox, T.}, \bibinfo{year}{2015}.
\newblock \bibinfo{title}{U-net: Convolutional networks for biomedical image
  segmentation}, in: \bibinfo{booktitle}{Int. Conf. on Medical
  Image Computing and Computer-Assisted Intervention},
  \bibinfo{organization}{Springer}, pp. \bibinfo{pages}{234--241}.
\bibitem[{Sadanandan et~al.(2017)Sadanandan, Ranefall, Le~Guyader and
  Wahlby}]{sadanandan17}
\bibinfo{author}{Sadanandan, S.K.}, \bibinfo{author}{Ranefall, P.},
  \bibinfo{author}{Le~Guyader, S.}, \bibinfo{author}{Wahlby, C.},
  \bibinfo{year}{2017}.
\newblock \bibinfo{title}{Automated training of deep convolutional neural
  networks for cell segmentation}.
\newblock \bibinfo{journal}{Scientific Reports} \bibinfo{volume}{7},
  \bibinfo{pages}{7860}.
\bibitem[{Sirinukunwattana et~al.(2016)Sirinukunwattana, Raza, Tsang, Snead,
  Cree and Rajpoot}]{sirinukunwattana16}
\bibinfo{author}{Sirinukunwattana, K.}, \bibinfo{author}{Raza, S.E.A.},
  \bibinfo{author}{Tsang, Y.W.}, \bibinfo{author}{Snead, D.R.},
  \bibinfo{author}{Cree, I.A.}, \bibinfo{author}{Rajpoot, N.M.},
  \bibinfo{year}{2016}.
\newblock \bibinfo{title}{Locality sensitive deep learning for detection and
  classification of nuclei in routine colon cancer histology images}.
\newblock \bibinfo{journal}{IEEE Transactions on Medical Imaging}
  \bibinfo{volume}{35}, \bibinfo{pages}{1196--1206}.
\bibitem[{Song et~al.(2017)Song, Tan, Jiang, Cheng, Ni, Chen, Lei and
  Wang}]{song17}
\bibinfo{author}{Song, Y.}, \bibinfo{author}{Tan, E.L.},
  \bibinfo{author}{Jiang, X.}, \bibinfo{author}{Cheng, J.Z.},
  \bibinfo{author}{Ni, D.}, \bibinfo{author}{Chen, S.}, \bibinfo{author}{Lei,
  B.}, \bibinfo{author}{Wang, T.}, \bibinfo{year}{2017}.
\newblock \bibinfo{title}{Accurate cervical cell segmentation from overlapping
  clumps in pap smear images}.
\newblock \bibinfo{journal}{IEEE Transactions on Medical Imaging}
  \bibinfo{volume}{36}, \bibinfo{pages}{288--300}.
\bibitem[{Song et~al.(2015)Song, Zhang, Chen, Ni, Lei and Wang}]{song15}
\bibinfo{author}{Song, Y.}, \bibinfo{author}{Zhang, L.}, \bibinfo{author}{Chen,
  S.}, \bibinfo{author}{Ni, D.}, \bibinfo{author}{Lei, B.},
  \bibinfo{author}{Wang, T.}, \bibinfo{year}{2015}.
\newblock \bibinfo{title}{Accurate segmentation of cervical cytoplasm and
  nuclei based on multiscale convolutional network and graph partitioning}.
\newblock \bibinfo{journal}{IEEE Transactions on Biomedical Engineering}
  \bibinfo{volume}{62}, \bibinfo{pages}{2421--2433}.
\bibitem[{Su et~al.(2015)Su, Xing, Kong, Xie, Zhang and Yang}]{su15}
\bibinfo{author}{Su, H.}, \bibinfo{author}{Xing, F.}, \bibinfo{author}{Kong,
  X.}, \bibinfo{author}{Xie, Y.}, \bibinfo{author}{Zhang, S.},
  \bibinfo{author}{Yang, L.}, \bibinfo{year}{2015}.
\newblock \bibinfo{title}{Robust cell detection and segmentation in
  histopathological images using sparse reconstruction and stacked denoising
  autoencoders}, in: \bibinfo{booktitle}{Int. Conf. on Medical
  Image Computing and Computer-Assisted Intervention},
  \bibinfo{organization}{Springer}, pp. \bibinfo{pages}{383--390}.
\bibitem[{Su et~al.(2013)Su, Yin, Huh and Kanade}]{su13}
\bibinfo{author}{Su, H.}, \bibinfo{author}{Yin, Z.}, \bibinfo{author}{Huh, S.},
  \bibinfo{author}{Kanade, T.}, \bibinfo{year}{2013}.
\newblock \bibinfo{title}{Cell segmentation in phase contrast microscopy images
  via semi-supervised classification over optics-related features}.
\newblock \bibinfo{journal}{Medical Image Analysis} \bibinfo{volume}{17},
  \bibinfo{pages}{746--765}.
\bibitem[{Xie et~al.(2018a)Xie, Noble and Zisserman}]{xie18a}
\bibinfo{author}{Xie, W.}, \bibinfo{author}{Noble, J.A.},
  \bibinfo{author}{Zisserman, A.}, \bibinfo{year}{2018}a.
\newblock \bibinfo{title}{Microscopy cell counting and detection with fully
  convolutional regression networks}.
\newblock \bibinfo{journal}{Computer Methods in Biomechanics and Biomedical
  Engineering: Imaging and Visualization} \bibinfo{volume}{6},
  \bibinfo{pages}{283--292}.
\bibitem[{Xie et~al.(2015a)Xie, Kong, Xing, Liu, Su and Yang}]{xie15b}
\bibinfo{author}{Xie, Y.}, \bibinfo{author}{Kong, X.}, \bibinfo{author}{Xing,
  F.}, \bibinfo{author}{Liu, F.}, \bibinfo{author}{Su, H.},
  \bibinfo{author}{Yang, L.}, \bibinfo{year}{2015}a.
\newblock \bibinfo{title}{Deep voting: A robust approach toward nucleus
  localization in microscopy images}, in: \bibinfo{booktitle}{Int.
  Conf. on Medical Image Computing and Computer-Assisted Intervention},
  \bibinfo{organization}{Springer}, pp. \bibinfo{pages}{374--382}.
\bibitem[{Xie et~al.(2015b)Xie, Xing, Kong, Su and Yang}]{xie15a}
\bibinfo{author}{Xie, Y.}, \bibinfo{author}{Xing, F.}, \bibinfo{author}{Kong,
  X.}, \bibinfo{author}{Su, H.}, \bibinfo{author}{Yang, L.},
  \bibinfo{year}{2015}b.
\newblock \bibinfo{title}{Beyond classification: Structured regression for
  robust cell detection using convolutional neural network}, in:
  \bibinfo{booktitle}{Int. Conf. on Medical Image Computing and
  Computer-Assisted Intervention}, \bibinfo{organization}{Springer}, pp.
  \bibinfo{pages}{358--365}.
\bibitem[{Xie et~al.(2018b)Xie, Xing, Shi, Kong, Su and Yang}]{xie18b}
\bibinfo{author}{Xie, Y.}, \bibinfo{author}{Xing, F.}, \bibinfo{author}{Shi,
  X.}, \bibinfo{author}{Kong, X.}, \bibinfo{author}{Su, H.},
  \bibinfo{author}{Yang, L.}, \bibinfo{year}{2018}b.
\newblock \bibinfo{title}{Efficient and robust cell detection: A structured
  regression approach}.
\newblock \bibinfo{journal}{Medical Image Analysis} \bibinfo{volume}{44},
  \bibinfo{pages}{245--254}.
\bibitem[{Xing et~al.(2014)Xing, Su, Neltner and Yang}]{xing14}
\bibinfo{author}{Xing, F.}, \bibinfo{author}{Su, H.}, \bibinfo{author}{Neltner,
  J.}, \bibinfo{author}{Yang, L.}, \bibinfo{year}{2014}.
\newblock \bibinfo{title}{Automatic Ki-67 counting using robust cell detection
  and online dictionary learning}.
\newblock \bibinfo{journal}{IEEE Transactions on Biomedical Engineering}
  \bibinfo{volume}{61}, \bibinfo{pages}{859--870}.
\bibitem[{Xu et~al.(2016)Xu, Xiang, Liu, Gilmore, Wu, Tang and
  Madabhushi}]{xu16}
\bibinfo{author}{Xu, J.}, \bibinfo{author}{Xiang, L.}, \bibinfo{author}{Liu,
  Q.}, \bibinfo{author}{Gilmore, H.}, \bibinfo{author}{Wu, J.},
  \bibinfo{author}{Tang, J.}, \bibinfo{author}{Madabhushi, A.},
  \bibinfo{year}{2016}.
\newblock \bibinfo{title}{Stacked sparse autoencoder (SSAE) for nuclei
  detection on breast cancer histopathology images}.
\newblock \bibinfo{journal}{IEEE Transactions on Medical Imaging}
  \bibinfo{volume}{35}, \bibinfo{pages}{119--130}.
\bibitem[{Yang et~al.(2006)Yang, Li and Zhou}]{yang06}
\bibinfo{author}{Yang, X.}, \bibinfo{author}{Li, H.}, \bibinfo{author}{Zhou,
  X.}, \bibinfo{year}{2006}.
\newblock \bibinfo{title}{Nuclei segmentation using marker-controlled
  watershed, tracking using mean-shift, and Kalman filter in time-lapse
  microscopy}.
\newblock \bibinfo{journal}{IEEE Transactions on Circuits and Systems I:
  Regular Papers} \bibinfo{volume}{53}, \bibinfo{pages}{2405--2414}.
\bibitem[{Zeiler(2012)}]{zeiler2012}
\bibinfo{author}{Zeiler, M.D.}, \bibinfo{year}{2012}.
\newblock \bibinfo{title}{AdaDelta: An adaptive learning rate method}.
\newblock \bibinfo{journal}{arXiv preprint arXiv:1212.5701} .


\end{thebibliography}


\end{document}